\documentclass[journal]{IEEEtran}
\usepackage{url}
\usepackage{graphicx}
\usepackage[numbers,sort&compress]{natbib}
\usepackage{subfigure}
\usepackage{threeparttable}
\usepackage{booktabs}
\renewcommand{\toprule}{\hline\hline}
\renewcommand{\midrule}{\hline}
\renewcommand{\bottomrule}{\hline\hline}
\renewcommand{\cmidrule}[1]{\cline{#1}}
\usepackage{multirow}
\usepackage{multicol}
\usepackage{amsmath}
\usepackage{multicol}
\usepackage{cuted}
\usepackage{hyperref}
\usepackage{threeparttable}
\begin{document}

\title{Rate-distortion Optimized Point Cloud Preprocessing for Geometry-based Point Cloud Compression}

\author{Wanhao Ma,
        ~Wei Zhang,~\IEEEmembership{Member,~IEEE,}
         ~Shuai Wan, ~\IEEEmembership{Member,~IEEE,}
        and~Fuzheng~Yang,~\IEEEmembership{Member,~IEEE}
\thanks{Wanhao Ma is with the School of Telecommunications Engineering, Xidian University, Xi'an, China (e-mail:whma24@stu.xidian.edu.cn).}
\thanks{Wei Zhang is with the School of Telecommunications Engineering, Xidian University, Xi'an, China and is also with Pengcheng Laboratory, Shenzhen, China (e-mail: wzhang@xidian.edu.cn).}
\thanks{Shuai Wan is with the School of Electronics and Information, Northwestern Polytechnical University, Xi’an, China, and also with the School of Engineering, Royal Melbourne Institute of Technology, Melbourne, VIC 3001, Australia (e-mail: swan@nwpu.edu.cn).}
\thanks{Fuzheng Yang is with the School of Telecommunications Engineering, Xidian University, Xi'an, China (e-mail: fzhyang@mail.xidian.edu.cn).}
\thanks{Wei Zhang is the corresponding author of this paper.}
}

\maketitle

\begin{abstract}
 Geometry-based point cloud compression (G-PCC), an international standard designed by MPEG, provides a generic framework for compressing diverse types of point clouds while ensuring interoperability across applications and devices. However, G-PCC underperforms compared to recent deep learning-based PCC methods despite its lower computational power consumption. To enhance the efficiency of G-PCC without sacrificing its interoperability or computational flexibility, we propose a novel preprocessing framework that integrates a compression-oriented voxelization network with a differentiable G-PCC surrogate model, jointly optimized in the training phase. The surrogate model mimics the rate-distortion behaviour of the non-differentiable G-PCC codec, enabling end-to-end gradient propagation. The versatile voxelization network adaptively transforms input point clouds using learning-based voxelization and effectively manipulates point clouds via global scaling, fine-grained pruning, and point-level editing for rate-distortion trade-offs. During inference, only the lightweight voxelization network is appended to the G-PCC encoder, requiring no modifications to the decoder, thus introducing no computational overhead for end users. Extensive experiments demonstrate a 38.84\% average BD-rate reduction over G-PCC. By bridging classical codecs with deep learning, this work offers a practical pathway to enhance legacy compression standards while preserving their backward compatibility, making it ideal for real-world deployment.
\end{abstract}


\IEEEpeerreviewmaketitle

\section{Introduction}
Point clouds, as a fundamental three-dimensional visual representation, find diverse applications in areas including autonomous driving~\cite{PointPillars, seg_pc,auto_driving_li}, virtual/augmented reality~\cite{pointxr, Immersive-Labeler,ar_han}, and 3D reconstruction \cite{3D-ARNet, yu2022part, rec_lin}. However, raw point clouds often exhibit significant data volume with redundancy, leading to high storage and bandwidth demands. This drives the need for point cloud compression (PCC) techniques. Depending on the compression target, PCC methods can be categorized into point cloud geometry compression (PCGC), which addresses redundancy in the geometry coordinates defining the structure of objects and scenes, and point cloud attribution compression (PCAC), which focuses on the compression of the point cloud appearance. In PCGC, significant advancements have been made with both traditional coding frameworks and learning-based compression pipelines. Notable examples include the G-PCC octree coding \cite{gpcc, octree}, a robust and broadly applicable framework standardized by the Moving Picture Experts Group (MPEG), and SparsePCGC \cite{sparsepcgc}, a learning-based codec that utilizes a convolutional representation of multiscale sparse tensors to achieve state-of-the-art geometry compression performance.

\begin{figure}[!t]
    \centering
    \includegraphics[width=\linewidth]{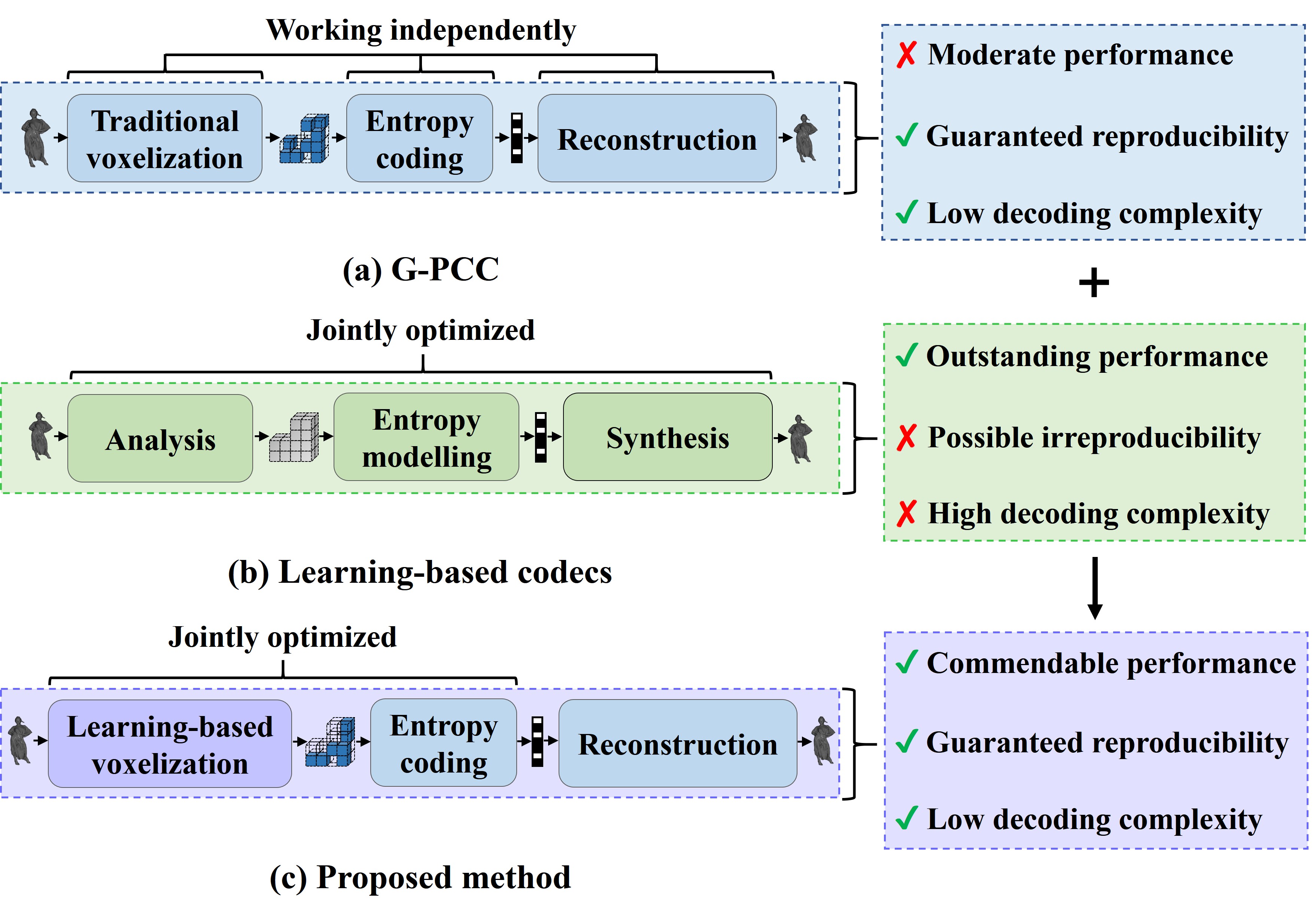}
    \caption{Advantages and disadvantages of different point cloud geometry compression pipelines. Our proposed method combines the advantages of G-PCC and learning-based codecs.}
    \label{fig:ad_disad}
\end{figure}

In general, learning-based point cloud codecs~\cite{ddpcc,pcgcv2,sparsepcgc,octattn, wang_lossy,huo_xiao}  leverage large datasets to model the point distribution pattern more accurately, achieving superior rate-distortion performance compared to traditional codecs. However, these methods are computationally intensive and rely on graphics processing units (GPUs) for inference, which is especially intolerable at the decoding side. Moreover, reproducibility remains a significant challenge for learning-based codecs~\cite{reproducible}, as the same trained neural network may produce inconsistent (or even undecodable~\cite{m70395}) results across different GPUs and software platforms such as PyTorch and TensorFlow. Ensuring consistent inference across platforms often requires either strict alignment of hardware and software or a significant compromise in compression performance, hindering the popularization of learning-based codecs. In contrast, G-PCC offers a generic framework for compressing diverse types of point clouds while guaranteeing reproducibility across applications and devices. However, despite its lower computational complexity, G-PCC lags behind recent deep learning-based PCC methods in compression performance.

Given these challenges, combining the strengths of G-PCC and learning-based methods is critical for the real-world deployment of PCC codecs. The superiority of learning-based methods lies in that all parts of the entire pipeline are jointly optimized, including analysis (e.g., preprocessing and latent representation), entropy modelling, and synthesis (e.g., latent feature decoding and enhancement). In contrast, the corresponding steps in G-PCC octree coding, namely voxelization (i.e., quantization and octree representation), entropy coding, and reconstruction (i.e., occupancy decoding and dequantization) work independently. We summarize the advantages and disadvantages of G-PCC and learning-based codecs in Fig.~\ref{fig:ad_disad}(a) and Fig.~\ref{fig:ad_disad}(b).

\begin{figure*}[!t]
    \centering
    \includegraphics[width=1\linewidth]{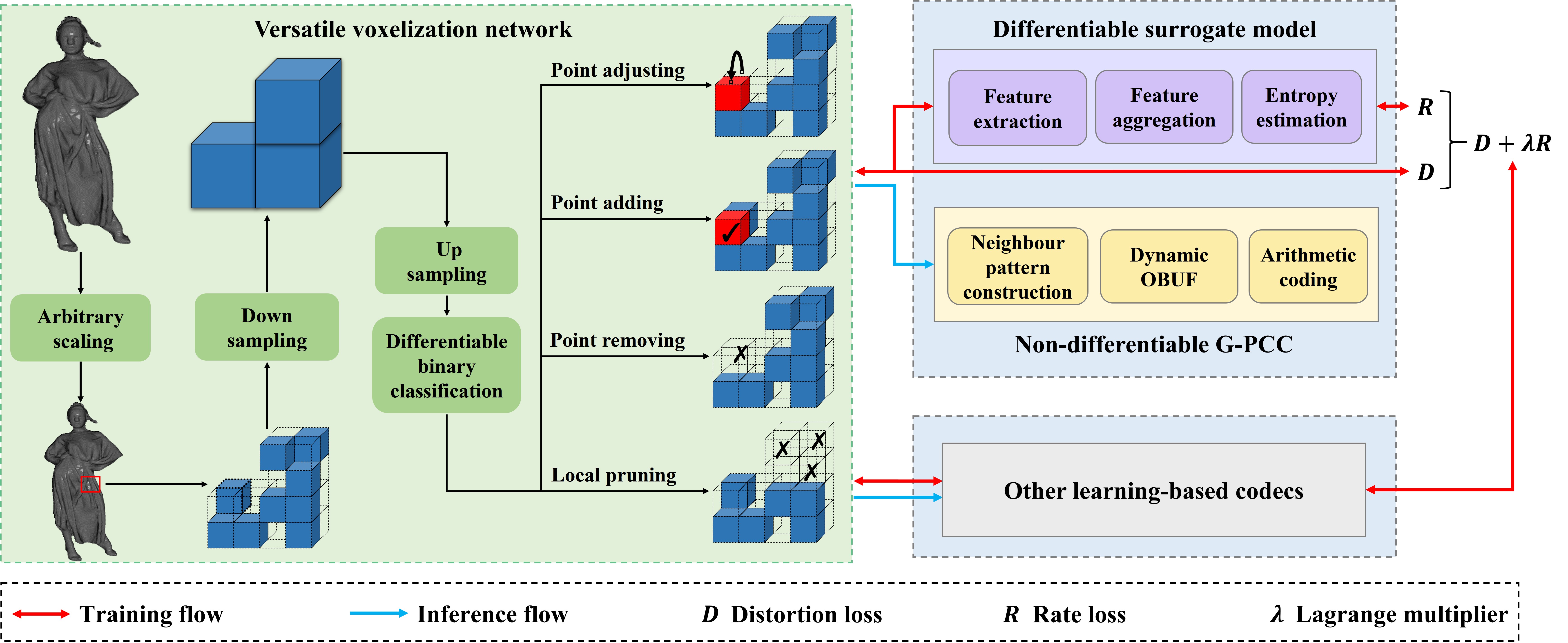}
    \caption{Overview of the proposed voxelization-compression joint optimization pipeline. The versatile voxelization network acts as a preprocessing module for G-PCC. Through joint optimization with a differentiable G-PCC surrogate model during training, the network learns to manipulate point clouds via global scaling, local pruning, and point editing (adding, removing, or adjusting) to optimize the rate-distortion tradeoff for G-PCC compression.}
    \label{fig:overview}
\end{figure*}

In this paper, we propose a practical solution that combines the advantages of G-PCC and learning-based methods as shown in Fig.~\ref{fig:ad_disad}(c). Specifically, we introduce a learning-based voxelization network followed by the G-PCC codec, enhancing G-PCC with guaranteed reproducibility but without additional computational burden on decoding. The voxelization network is jointly optimized with the proposed differentiable G-PCC surrogate during training, enabling compression-oriented voxelization. The network learns to optimize the point cloud distribution by global scaling, local pruning, and point-level editing, ensuring that the processed point cloud can be compressed effectively by the G-PCC codec at optimal rate-distortion trade-offs. A dedicated back-loaded upsampling design, which can significantly reduce floating-point operations (FLOPs), is proposed to lighten the voxelization network. Importantly, this solution avoids the need for learned networks at the decoding side, thus ensuring computational efficiency and reproducibility across platforms. An overview of the proposed pipeline is shown in Fig.~\ref{fig:overview}.

The contributions of this paper are summarized as follows:
\begin{itemize}
    \item We propose a lightweight, versatile point cloud voxelization network for preprocessing. When combined with G-PCC, this network can manipulate the point cloud at various granularities, ensuring the processed point cloud is compressed by the G-PCC codec at optimal rate-distortion tradeoffs. 
    \item We propose a differentiable G-PCC surrogate model that mimics the behaviour of G-PCC. This surrogate model can replace the non-differentiable G-PCC in gradient-based optimization, enabling joint optimizations with any point cloud-oriented tasks requiring the compression module. Furthermore, when used alone, the surrogate model serves as a lossless PCC codec, achieving state-of-the-art performance.
    \item We present a practical paradigm that offers a deployable solution to significantly enhance traditional G-PCC standards. It features a user-friendly design that requires no modifications at the decoding side and introduces no additional computational burden for end users. This approach maintains backward compatibility, reproducibility, and computational efficiency, providing a clear path toward real-world adoption.
\end{itemize}

\section{Related Work}
This section provides a brief introduction and discussion of related research in point cloud geometry preprocessing, G-PCC geometry compression, and learning-based geometry compression.

\subsection{Point Cloud Geometry Preprocessing}
We classify geometry preprocessing methods in the literature into general-purpose methods, which are not specific to a particular downstream task, and compression-oriented methods. The general-purpose methods are usually point-based and process the point cloud for denoising, simplification, and densification. For point cloud denoising, Alexa et al.~\cite{point_cloud_surface} propose to approximate the point cloud with a smooth surface using the moving least squares (MLS) method. Rakotosaona et al.~\cite{pointcleannet} decompose the denoising task into two subtasks: outlier removal and point position adjustment, training two neural networks separately using noise-free point clouds as ground truth. For point cloud simplification, farthest point sampling (FPS)~\cite{fps_1, fps_2} is the most widely used method, which selects the farthest points iteratively to reduce density. For point cloud densification, Yu et al.~\cite{punet} construct pairs of low-density and high-density point clouds to supervise the training of the neural network, using a joint objective function to ensure uniform density on the underlying surface. Beyond supervised training, Li et al.\cite{pugan} design a GAN-based network with a compound loss to enhance output point distribution uniformity.

Compression-oriented preprocessing methods are relatively scarce. In classical codecs, G-PCC employs conventional voxelization~\cite{voxelization_hinks}, which involves quantizing point cloud coordinates before octree construction and occupancy coding. While this approach reduces the resolution of geometry coordinates for compression, it often introduces noticeable distortion in reconstructed point clouds~\cite{dingquan_sr} and lacks optimal synergy between preprocessing and compression. In learning-based codecs, Wang et al.~\cite{pcgcv2} downsample voxel-based point clouds before encoding to reduce bit costs. Huang et al. \cite{point_based_huang} use FPS to identify local region centres and extract features from their $K$-nearest neighbours for entropy modelling. However, none of these methods are jointly optimized with the compression task, leaving significant room for improvement in achieving better rate-distortion performance.

\subsection{Compression-oriented Image Preprocessing}
Compression-oriented image preprocessing converts the image into a form more suitable for the rate-distortion characteristics of codecs. Talebi et al. \cite{preediting} propose an image pre-editing method for JPEG \cite{jpeg}. They design a learned convolutional neural network and train the network jointly with a differentiable JPEG surrogate. Son et al. \cite{cpnet} propose a compact representation network to reduce the capacity of an input image, which is jointly optimized with a JPEG-based auxiliary codec network and a bit estimation network. Unlike above JPEG-oriented methods, Yang et al. \cite{preprocessor} propose an image processor that is jointly optimized with a learning-based differentiable codec without the need for a surrogate model. 

\subsection{G-PCC Octree Geometry Coding}
G-PCC~\cite{gpcc}, developed by the MPEG, is a state-of-the-art international standard for PCC. The 1st Edition was published in 2023, with an enhanced version, called~\textit{E-G-PCC}, slated for release in 2026. G-PCC tackles the challenge of compressing diverse geometries through three coding modes: octree coding~\cite{octree}, trisoup coding~\cite{gpcc_trisoup}, and predictive tree coding~\cite{gpcc_pred}. Trisoup coding is optimized for solid and dense point clouds where explicit surface reconstruction is required. Predictive tree coding is tailored for spinning LiDAR point clouds. Octree coding, on the other hand, serves as a versatile tool applicable to all types of point clouds.

In octree-based coding, a hierarchical octree structure is created by recursively subdividing the 3D space occupied by the point cloud into eight smaller regions (octants). This approach enables a compact and scalable representation of the point cloud geometry. The octree structure converts the geometry coordinates to a cascade of binary octree node occupancy bits arranged in breadth-first order. These bits are then arithmetically coded with contextual information. The coding process can be divided into three main steps: neighbourhood pattern construction, dynamic Optimal Binary coder with Update on the Fly (dynamic OBUF)~\cite{obuf}, and arithmetic coding. In the first step, contextual information, termed as~\textit{neighbour pattern}, is derived by examining specific parent neighbours and already-coded sibling nodes of the current node. To manage the exponential growth of contexts as the number of neighbours increases, dynamic OBUF is employed. This tool maps numerous neighbour configurations to a smaller set of dynamically updated context probability models (CPMs), effectively reducing context dimensionality. Finally, the occupancy bit is arithmetically encoded based on the mapped CPM. 

Apart from the above main octree coding process, G-PCC octree coding also incorporates specialized tools to address specific scenarios. For point clouds with planar structures, planar mode~\cite{gpcc_planar} activates a dedicated context model to code the planar information. When dealing with sparse nodes containing isolated points, direct coding mode~\cite{gpcc_icm} encodes the relative positions of the points, avoiding unnecessary octree subdivisions. These supplementary tools target edge cases in point cloud distributions, further boosting the overall coding efficiency of G-PCC.

\subsection{Learning-based Point Cloud Geometry Coding}
While G-PCC leads in traditional PCC methods, it underperforms recent deep learning-based PCC frameworks. Existing learning-based geometry codecs \cite{voxel_based_Quach,scp,pcgcv2,wu_prior,Nguyen} typically consist of three components: data representation, entropy coding, and reconstruction. First, data representation transforms the point clouds into intermediate representations (e.g., latent codes). Second, entropy coding exploits context information in intermediate representations and encodes them efficiently. Finally, the reconstruction module converts the decoded intermediate representations back into point clouds. Quach et al. \cite{voxel_based_Quach} propose to use 3D convolutional layers to convert point clouds to latent codes and back, adopting the factorized entropy model~\cite{balle_end_to_end} for entropy coding. PCGCv2 \cite{pcgcv2} introduces 3D sparse convolutions that only operate on non-empty voxels. It downsamples point clouds before converting them into latent codes and upsamples them after transforming decoded latent codes back into point clouds. In~\cite{unicorn}, Wang et al. propose to convert point clouds into two sets of intermediate representations: downsampled point clouds and latent codes. The latent codes act as sub-band information to aid upsampling during reconstruction. SCP \cite{scp} transforms LiDAR point clouds from Cartesian coordinates to Spherical coordinates, which are more suitable for entropy coding.

It is noted that the data representation, entropy coding and reconstruction modules in learning-based compression models are jointly optimized in the training phase, enabling superior performance over traditional methods. However, this advantage comes at the cost of increased model complexity and computational demands. A recent review highlights a critical limitation: learning-based codecs suffer from reproducibility challenges~\cite{reproducible,m70395}, where hardware and software dependencies during implementation introduce inconsistencies in compression outcomes. Resolving this issue requires either strict alignment of hardware and software, or at the cost of significant compromise in compression performance. To enable practical deployment, hybrid approaches that integrate the robustness of traditional codecs with the adaptability of learning-based techniques may offer a viable path forward, particularly for applications requiring cross-platform consistency.

\begin{figure*}[!t]
    \centering
    \includegraphics[width=1\linewidth]{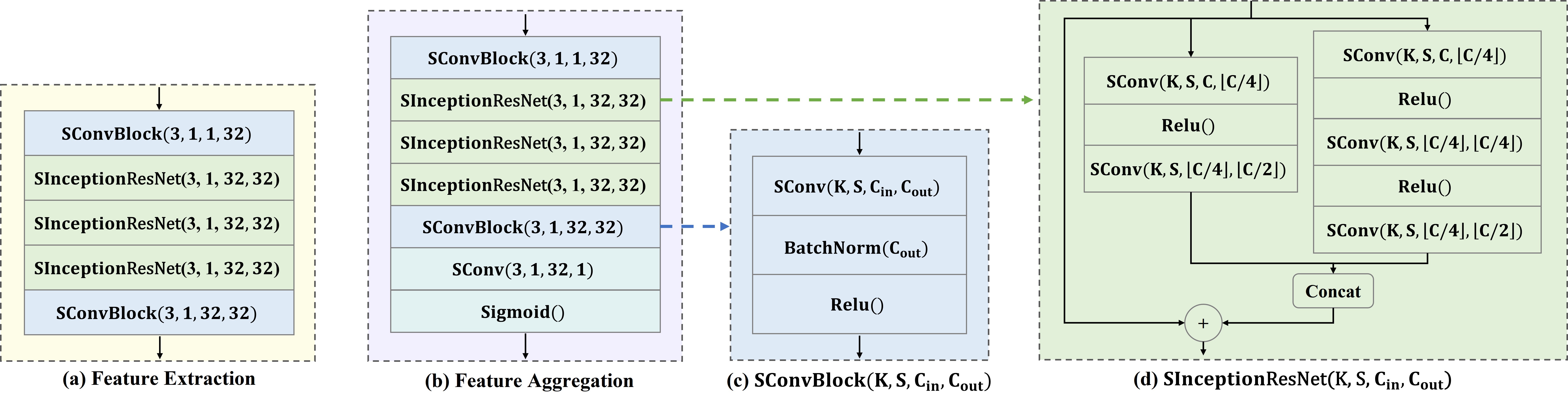}
    \caption{Some basic blocks of the proposed network architecture. Feature extraction (a) and feature aggregation (b) are basic modules of our method, which are composed of SConvBlock (c) and SInceptionResNet (d). We combine an SConv module, a BatchNorm module, and a Relu module into the SConvBlock. Sparse InceptionResNet, i.e., SInceptionResNet, is built by applying sparse convolution to the InceptionResNet \cite{inception_resnet}.}
    \label{fig:basic_blocks}
\end{figure*}

\section{Proposed Architecture}
\subsection{Overview}
This section introduces the proposed hybrid framework, which integrates a learning-based voxelization network with the traditional G-PCC codec. To enable joint optimization, we construct a differentiable surrogate model, which acts as a proxy of the non-differentiable G-PCC during training. Both the voxelization and surrogate networks adopt the multiscale sparse tensor representation of the point cloud from~\cite{sparsepcgc}, as it enables the one-to-one mapping to the octree structure used in G-PCC. Specifically, modifying the occupancy code in the sparse tensor directly corresponds to manipulating nodes in the octree hierarchy, ensuring alignment between the learned and traditional coding processes. Figure~\ref{fig:overview} provides an overview of the proposed framework. The differentiable surrogate model simulates the rate-distortion behaviours of the non-differentiable G-PCC, enabling our versatile voxelization network to be jointly optimized with G-PCC. The proposed voxelization network acts as a preprocessing module for G-PCC, manipulating point clouds via global scaling, local pruning, and point editing. For clarity, the core modules of the proposed network architecture are assembled into basic functional units, as shown in Fig.~\ref{fig:basic_blocks}.

\begin{figure}[!t]
    \centering
    \subfigure[\scriptsize Existing Upsampling Module]{
        \includegraphics[width=0.4\linewidth]{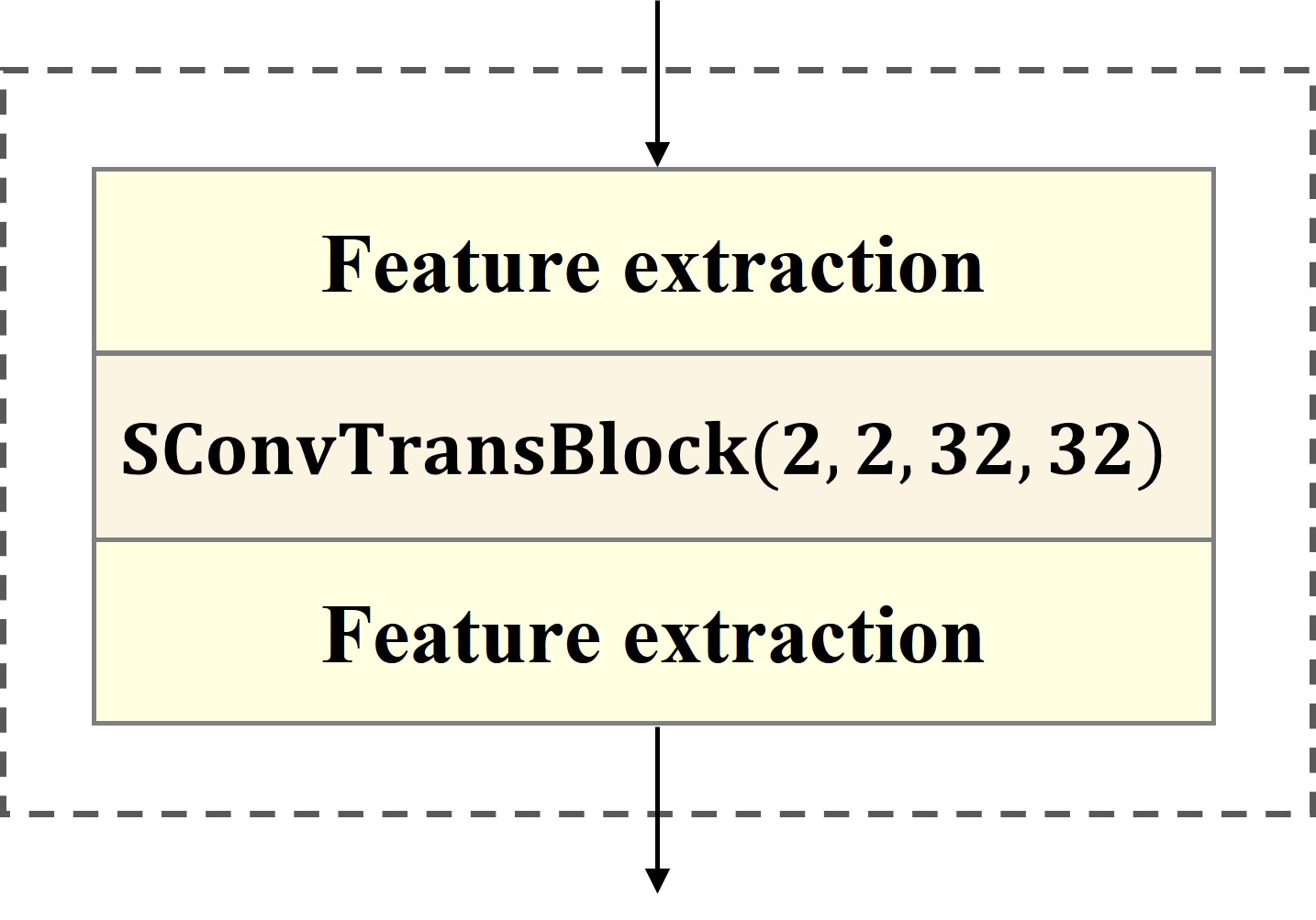}
        \label{fig:pre_upsample}
    }
    \subfigure[\scriptsize Back-loaded Upsampling Module]{
        \includegraphics[width=0.4\linewidth]{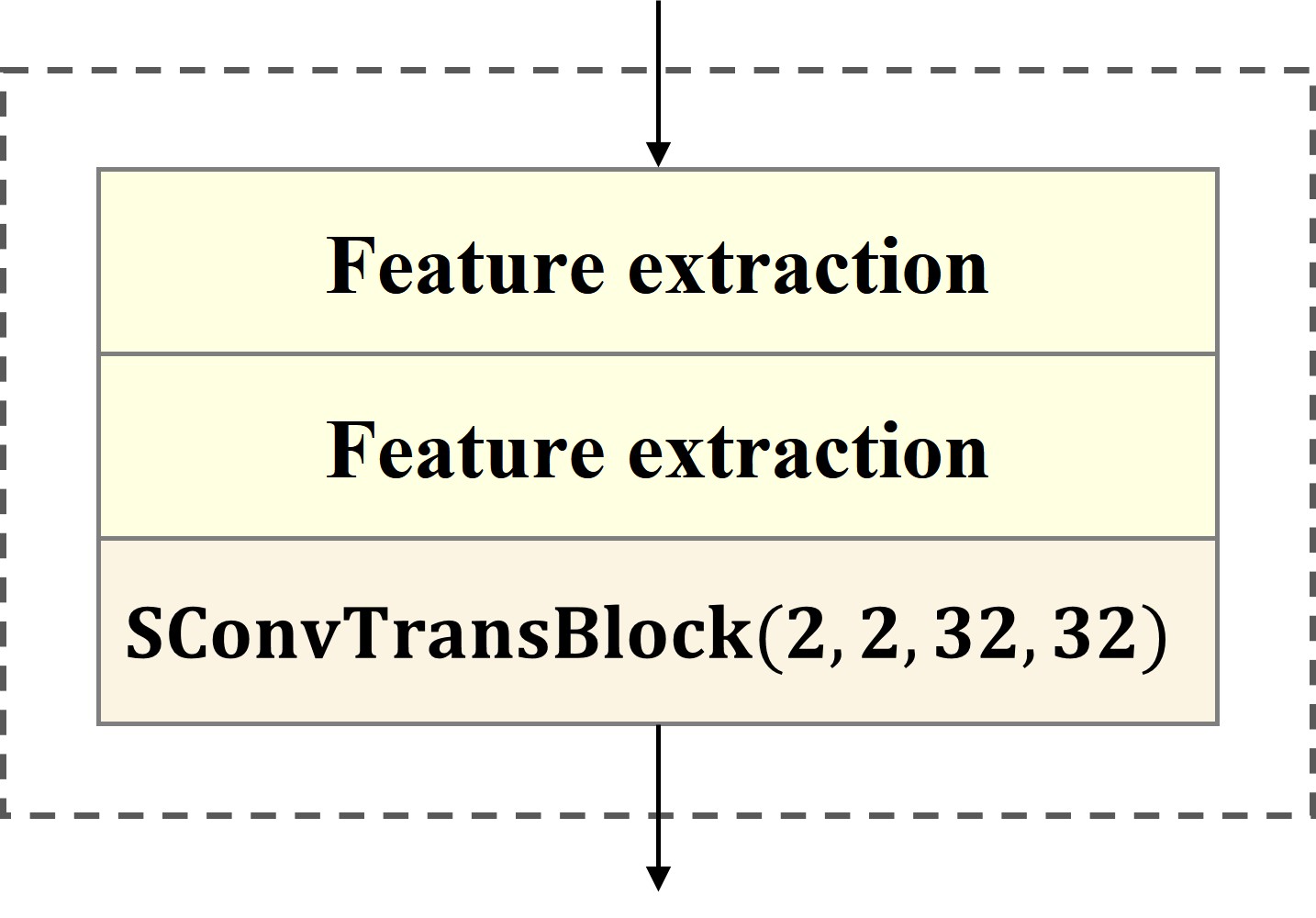}
        \label{fig:post_upsample}
   }
    \caption{Comparison of the existing upsampling module and the proposed back-loaded upsampling module. Positioning the transposed sparse convolution (i.e., the upsampling module) to the end can significantly reduce computational complexity. }
    \label{fig:different_upsample}
\end{figure}

\subsection{Versatile Voxelization Network}
The proposed versatile voxelization network aims to manipulate the point cloud with different granularity via global scaling, local pruning, and point-level adjusting such that the processed point cloud can be compressed by the G-PCC codec at optimal rate-distortion tradeoffs. The network is built on sparse convolutions~\cite{me} with the point cloud represented as multiscale sparse tensors.

The network supports arbitrary point cloud geometry precision by first scaling the input point at different quantization steps. Then, to locally adjust the point distribution, the scaled point cloud is edited by a classification mechanism that classifies the leaf nodes of the scaled point cloud as occupied or unoccupied based on the parent-child spatial correlations. Specifically, a downsampling block, implemented as a sparse convolution layer with a kernel size of 2 and a stride of 2, downsamples the sparse tensor by a factor of 2, transitioning the leaf nodes to their parent nodes in the octree. The downsampled sparse tensor is then upsampled by a back-loaded upsampling module, as shown in Fig.~\ref{fig:different_upsample} (b), which includes a series of feature extraction modules followed by a back-loaded transposed sparse convolution block, mapping the sparse tensor back to the child node layer thus achieving the upsampling. A detailed analysis of the back-loaded upsampling module is provided in Sec.~\ref{sec:post}. 

\begin{figure*}[!t]
    \centering
    \includegraphics[width=0.9\linewidth]{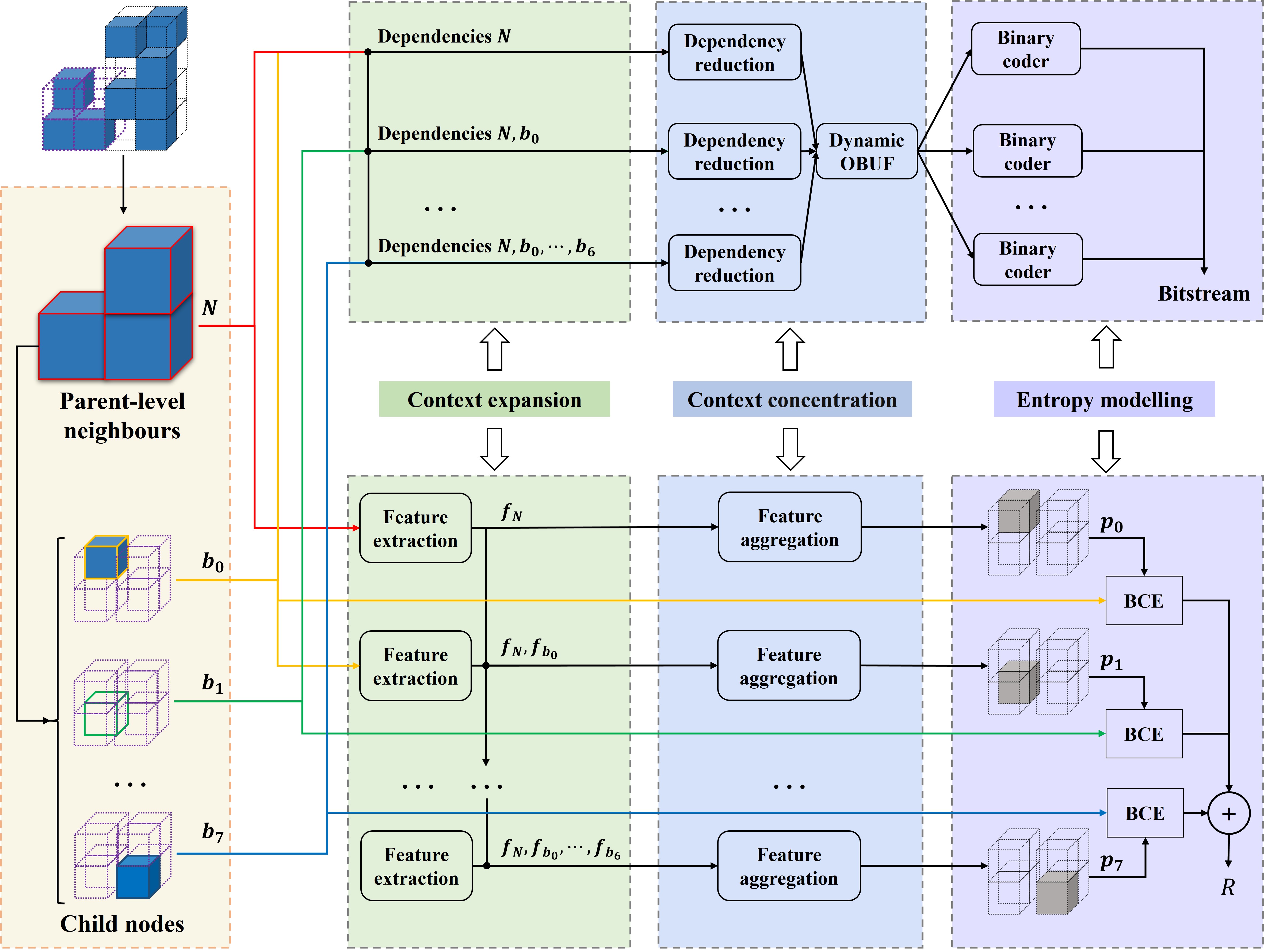}
    \caption{The workflow comparison of the non-differentiable G-PCC (top half) and the differentiable surrogate model (bottom half). The BCE is the abbreviation for binary cross-entropy.}
    \label{fig:gpcc_aux_comp}
\end{figure*}

Finally, the hidden features are binary classified. They pass through a Sigmoid function to compute the occupancy probabilities $p^{c}$ for all nodes in the child layer. Based on these probabilities, the voxelization network classifies the node as occupied if its occupancy probability $p^{c}$ exceeds 0.5, and as unoccupied otherwise. To achieve joint optimization with the proposed differentiable G-PCC surrogate, STERound function~\cite{ste} is employed to enable gradients to be propagated. The forward pass of STERound is described as 
\begin{equation}
        S(p^{c}) = 
            \left\{ \begin{array}{l}
                1, ~ \text{if} ~ p^{c} \geq 0.5 \\
                0, ~ \text{if} ~ p^{c} < 0.5 
            \end{array},\right.
    \label{equ:ste_round}
\end{equation}
and the backward pass as
\begin{equation}
    \frac{\partial S(p^{c})}{ \partial p^{c}} = 1,
\end{equation}
where $S$ represents the STERound operator.

By applying the aforementioned differentiable design, the versatile voxelization network dynamically adjusts the distribution of leaf nodes, enabling point adding (i.e., $p^c>0.5$), point removing (i.e., $p^c<0.5$), and octree pruning (i.e., $\quad p^c_i < 0.5,\forall i \in R$ where $R$ is the pruned region).

It is worth noting that the proposed versatile voxelization network is not limited to joint optimization with traditional G-PCC. It can also be seamlessly integrated into learning-based PCC frameworks. Additionally, it can serve as a point cloud manipulating module for other point cloud processing tasks such as classification and registration. This flexibility highlights the broader applicability of our approach beyond compression.

\subsection{Back-Loaded Upsampling Module}
\label{sec:post} 
Compared with the existing upsampling process in sparse tensor networks~~\cite{pcgcv2,sparsepcgc,zhang2023scalable,lodhi2023sparse,yim2024mamba} which positions the upsampling operation as an intermediate module among the feature extraction convolutional layers, we propose an upsampling module which back-loaded the upsampling operation to reduce the number of FLOPs while maintaining performance, leading to a lightweight voxelization network. A clear illustration of the proposed structure against the existing module is depicted in Fig.~\ref{fig:different_upsample}.

In neural networks, due to the diversity of operators, precisely counting FLOPS is complex. For simplicity, existing works~\cite{wang2024yolov9,Kong_2021_CVPR,tan2019efficientnet, Zhang_2018_CVPR,Ma_2024_CVPR, howard2017mobilenets, Sandler_2018_CVPR} approximate FLOPs as twice the number of multiplication operations. Adopting this convention, the number of FLOPs, denoted as $F$, for a single sparse convolution can be expressed as:
\begin{equation}
    F=2 N_a C_{i} C_{o},
\end{equation}
where $N_a$ represents the number of activated coordinates~\cite{me} in the input sparse tensor, and $C_i$ and $C_o$ denote the number of input and output channels, respectively. In contrast to existing upsampling modules, the back-loaded upsampling operation is placed at the tail end and thus does not alter $C_i$ and $C_o$. Consequently, $F$ is linear with $N_a$, i.e., $O(F)=O(N_a)$. If the upsampling operation occurs in the middle of the network, the additional voxels generated from upsampling will participate in subsequent operations, leading to increased $N_a$ and, consequently, $F$. A performance comparison is conducted between the existing upsampling module and the proposed back-loaded upsampling module. When integrated with the existing voxelization network, the back-loaded structure incurs only a slight 1.02\% BD-rate loss but an 85.50\% FLOPs reduction, demonstrating its ability to significantly reduce FLOPs while minimizing performance degradation.

\subsection{Differentiable G-PCC Surrogate} \label{sec:aux}
The differentiable G-PCC surrogate model aims to mimic the coding behaviours of G-PCC \cite{gpcc}. When jointly training the compression-oriented versatile voxelization network together with G-PCC, it replaces the non-differentiable G-PCC to enable backward propagation. 

The design of the G-PCC surrogate aligns with the octree coding pipeline in G-PCC. Figure.~\ref{fig:gpcc_aux_comp} illustrates the detailed modules of the proposed G-PCC surrogate and its correspondence with each G-PCC octree coding module. According to the codec description~\cite{gpcc_codec_disc} of G-PCC, G-PCC octree coding can be mainly divided into three stages: context expansion, context concentration, and entropy modelling. Specifically, as child nodes $b_i$ are encoded one by one, the context information for coding $b_i$ expands from parent-level neighbour configuration $N$ to ($N,b_0,b_1,\ldots,b_{i-1}$) considering previously-coded siblings. Then, to fully use the neighbour information while reducing the memory footprint of the obtained context information, Dynamic OBUF adaptively maps various contexts to a compact representation. Finally, $b_i$ is contextually encoded by the binary coder~\cite{cabac}. In essence, G-PCC is an autoregressive approach that uses previously-encoded nodes as conditions when coding the current node. It decomposes the probability of an 8-bit occupancy byte into eight conditional probabilities corresponding to each child node, converting an 8-bit non-binary coder into a cascade of eight binary coders, as described by Eq.(~\ref{equ:gpcc_compact}):
\begin{equation}
    H(b_0, b_1, \ldots, b_7 | N) = H(b_0| N) \prod_{i=1}^{7}H(b_i | N, b_0, \ldots, b_{i-1}).
\label{equ:gpcc_compact}
\end{equation}

To mimic G-PCC, the proposed G-PCC surrogate network also contains the above three components with the same functionality but replaces those non-differentiable operations with their corresponding differentiable alternatives. For context expansion, before encoding the first child node, the feature extraction block, as shown in Fig.~\ref{fig:basic_blocks}(a), extracts hidden features $f_N$ from nodes at the parent level, replacing the hand-crafted neighbourhood pattern configuration $N$ in G-PCC. As child nodes ${b_i}$ are encoded one by one, the extracted hidden features ($f_N,f_{b_0},f_{b_1},\ldots,f_{b_i}$) aggregate gradually. For context concentration, the feature aggregation block, as shown in Fig.~\ref{fig:basic_blocks}(b), is designed to achieve the aggregation of hidden features and outputs the occupancy probability of the node. As such, the surrogate network, represented as Eq.~(\ref{equ:aux_net}), is a differentiable relaxation of G-PCC: 
\begin{equation}
  H(b_0,b_1,\ldots, b_7|f_N) 
 = H(b_0|f_N) \prod_{i=1}^{7} H(b_i|f_N,f_{b_0},\ldots, f_{b_{i-1}}),
\label{equ:aux_net}
\end{equation} 
where $f_{b_i}$ corresponds to the feature of $b_i$.
It can be seen that Eq.~\ref{equ:gpcc_compact} and Eq.~\ref{equ:aux_net} are the same in their format, indicating that the proposed surrogate can be used as a proxy for G-PCC. The difference between them is that the discrete dependencies $(N,b_0,\ldots,b_i)$ used by non-differentiable G-PCC are changed into continuous hidden features $(f_N, f_{b_0},\ldots, f_{b_i})$ generated from neural networks.  

It should be noted that in addition to serving as the G-PCC surrogate for joint optimization, the proposed network can also be used alone as a lossless point cloud geometry codec.

\subsection{Loss Function}
To better introduce the loss function, we denote $N_s$ and $N_v$ as the number of points in the scaled point cloud and the voxelized point cloud, respectively, and $B$ as the BCE loss described in Eq.~(\ref{equ:bce}), i.e.,
\begin{equation}
    B(b,p)=-b \cdot log(p)-(1-b) \cdot log(1-p),
    \label{equ:bce}
\end{equation}
where $b$ represents occupancy bit, and $p$ represents occupancy probability.

For pretraining the surrogate model, we calculate the BCE loss between the occupancy bit $b_j^{v}$ of the voxelized point cloud and its estimated occupancy probability $p_j^{s}$ from our surrogate model. The  loss function for pretraining is as follows:
\begin{equation}
    L_{p}=\sum_{j=1}^{N_v} B(b_j^{v},p_j^{s}).
    \label{equ:pre_loss}
\end{equation}.

For joint optimization of the proposed versatile voxelization network and the G-PCC surrogate model, we use $L=D+\lambda R$ as the loss function~\cite{balle_end_to_end}, where $D$ represents distortion and $R$ represents rate. For distortion $D$, we calculate the binary cross entropy (BCE) loss between the occupancy bit $b_i^s$ of the scaled point cloud and its corresponding classification probability $p_i^{c}$ in our voxelization network. For rate $R$, we estimate the average coding length through its estimated occupancy probability $p_j^{s}$ generated from our surrogate model with fixed parameters. In total, our joint loss function for joint training is as follows:
\begin{equation}
    L=\sum_{i=1}^{N_s} B(b_i^s,p_i^c)+\lambda \sum_{j=1}^{N_v}B(b_j^v,p_j^{s}).
    \label{equ:obj}
\end{equation}

It's important to note that the BCE loss can serve different purposes based on different objects. Specifically, BCE can be used to measure classification accuracy, indicating distortion, or to estimate the average coding length, which represents rate. In this case, the first BCE component is utilized to assess the distortion between the original point cloud and the preprocessed point cloud. Meanwhile, the second BCE component is used to estimate the average coding length of the preprocessed point cloud in conjunction with the surrogate model.

\begin{figure}
    \centering
    \includegraphics[width=0.8\linewidth]{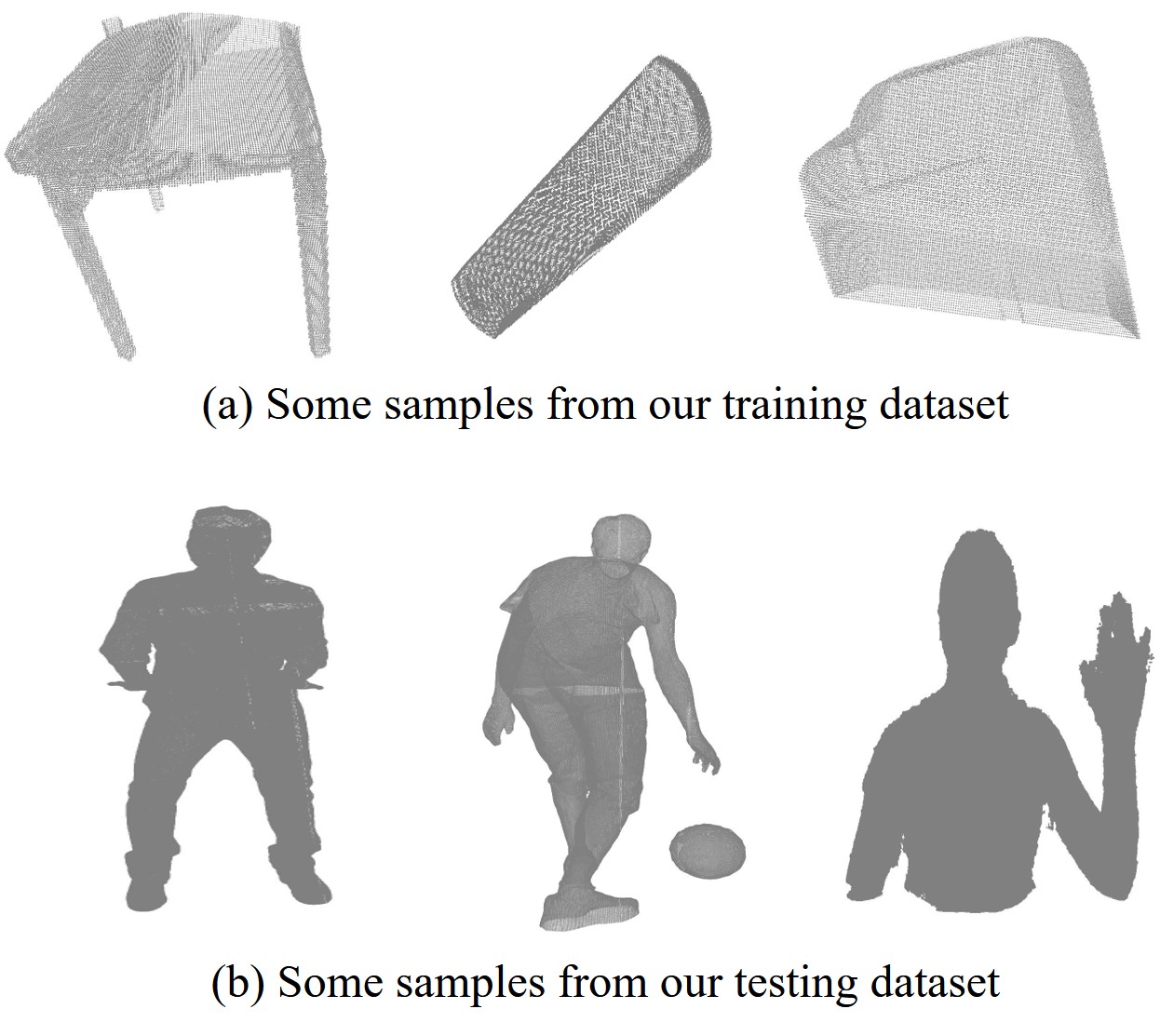}
    \caption{Visualizations of some samples in our training and testing datasets.}
    \label{fig:data_vis}
\end{figure}

\begin{table}[t!]
    \centering
    \caption{Performance gains achieved by integrating the proposed voxelization network as a preprocessing module for G-PCC (TMC13-v23) and SparsePCGC.}
    \label{tab:bd_rate}
    \renewcommand{\arraystretch}{1.1}
    \begin{tabular}{cc|c|c|c|c}
    \hline
    \hline
    \multirow{2}[2]{*}{Dataset} & \multirow{2}[2]{*}{Point Cloud} & \multicolumn{2}{c|}{G-PCC \cite{gpcc}} & \multicolumn{2}{c}{SparsePCGC \cite{sparsepcgc}} \\
      &   & \multicolumn{1}{c}{D1} & D2 & \multicolumn{1}{c}{D1} & D2 \\
    \hline
    \multirow{4}[2]{*}{Owlii} & basketball\_player & -43.72\% & -34.56\% & -2.77\% & -3.13\% \\
      & dancer & -41.34\% & -28.23\% & -2.29\% & -2.60\% \\
      & exercise & -41.77\% & -28.93\% & -3.63\% & -4.42\% \\
      & model & -39.70\% & -24.51\% & -2.45\% & -4.69\% \\
    \hline
    \multirow{4}[2]{*}{8iVFB} & longdress & -42.12\% & -26.54\% & -2.45\% & -2.95\% \\
      & loot & -42.58\% & -28.87\% & -2.48\% & -2.93\% \\
      & redandblack & -35.89\% & -20.26\% & -1.45\% & -1.68\% \\
      & soldier & -34.72\% & -19.45\% & -2.16\% & -2.43\% \\
    \hline
    \multirow{4}[2]{*}{MVUB} & andrew & -35.99\% & -30.49\% & -1.42\% & -2.12\% \\
      & david & -37.65\% & -35.61\% & -1.36\% & -2.13\% \\
      & phil & -37.46\% & -33.93\% & -0.31\% & -0.77\% \\
      & sarah & -38.43\% & -32.99\% & -1.87\% & -2.80\% \\
    \hline
      & queen & -33.59\% & -18.31\% & -4.44\% & -5.06\% \\
    \hline
      & Average & -38.84\% & -27.90\% & -2.24\% & -2.90\% \\
    \hline
    \hline
    \end{tabular}%
\end{table}

\begin{figure*}[!t]
    \centering
    \subfigure[D1 PSNR]{
        \includegraphics[width=0.25\linewidth]{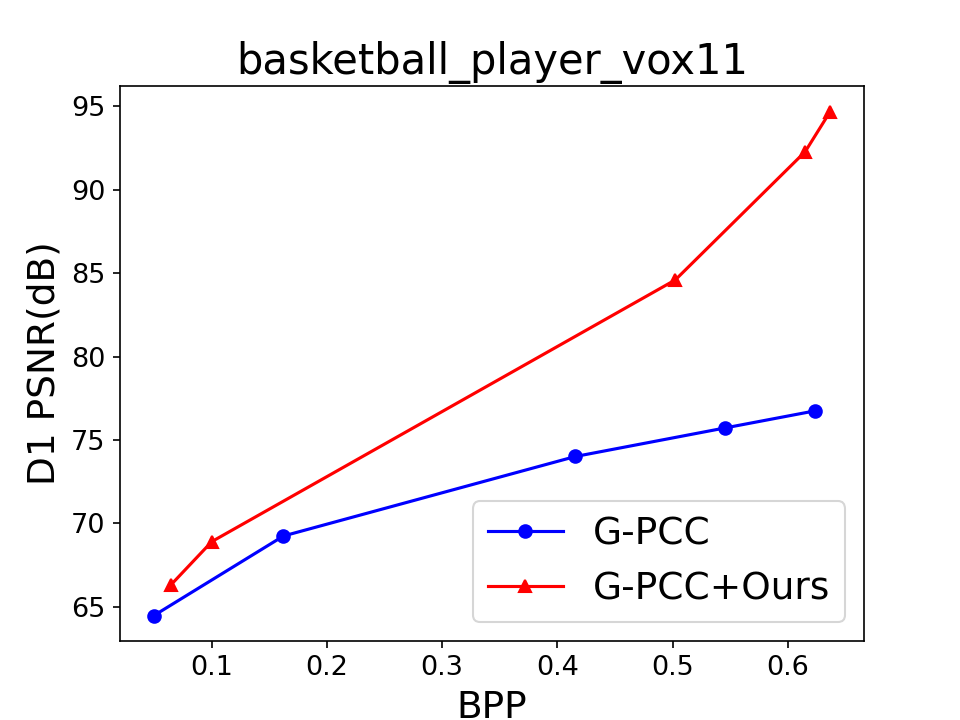}
        \includegraphics[width=0.25\linewidth]{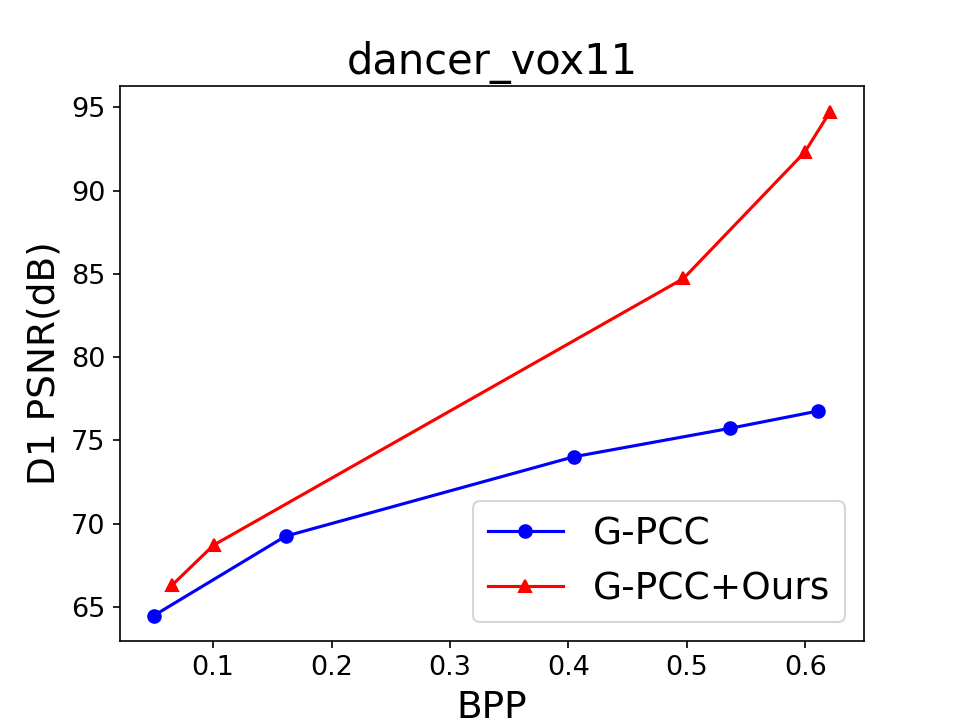}
        \includegraphics[width=0.25\linewidth]{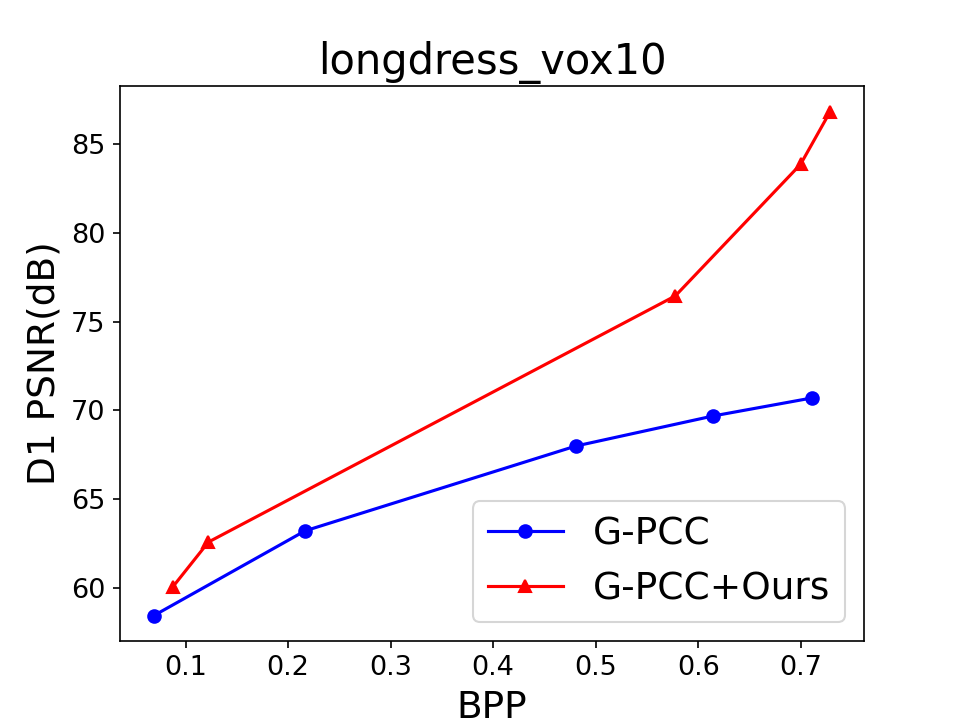}
        \includegraphics[width=0.25\linewidth]{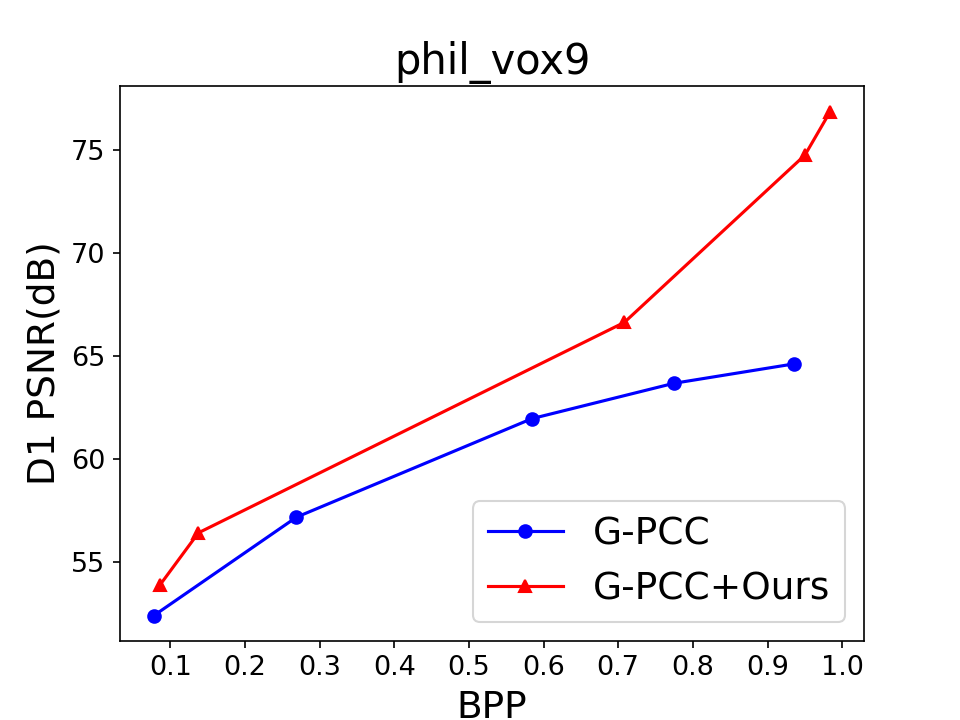}
    }
    \subfigure[D2 PSNR]{
        \includegraphics[width=0.25\linewidth]{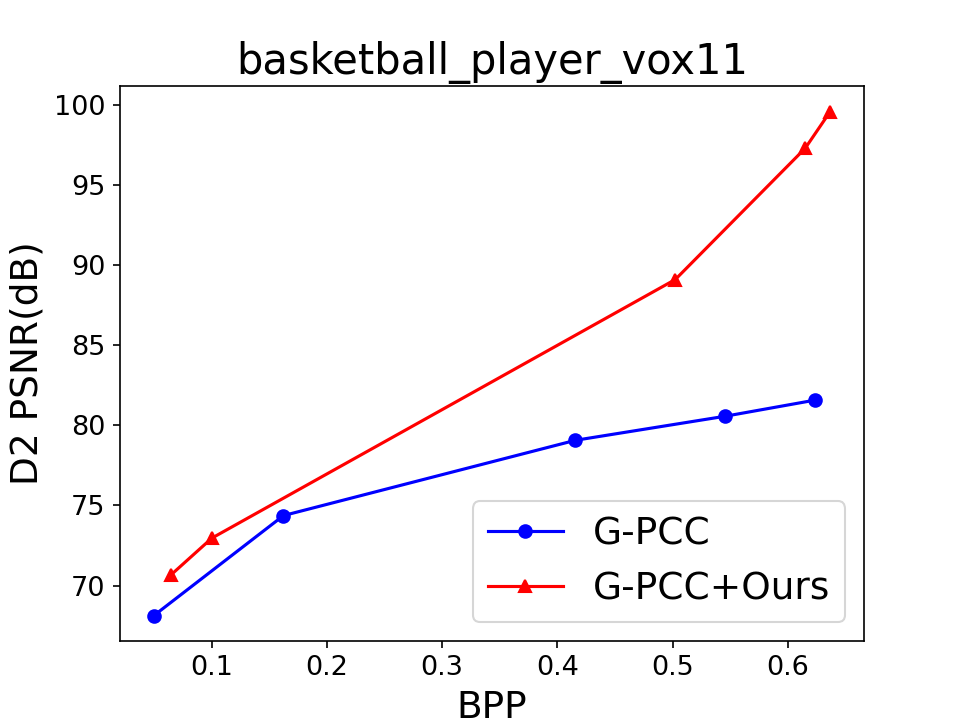}
        \includegraphics[width=0.25\linewidth]{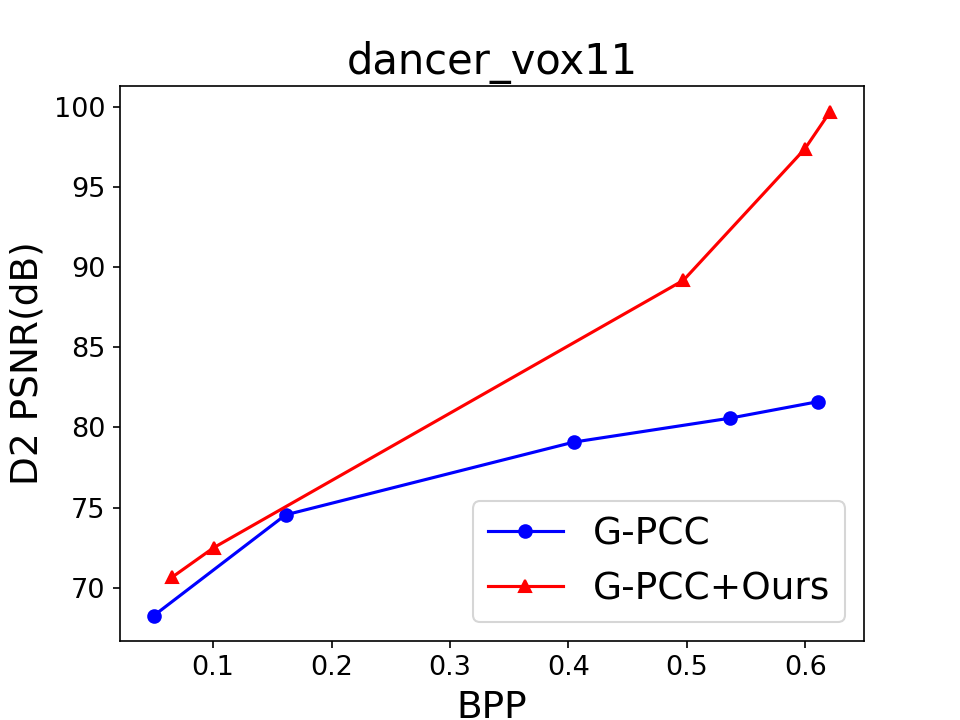}
        \includegraphics[width=0.25\linewidth]{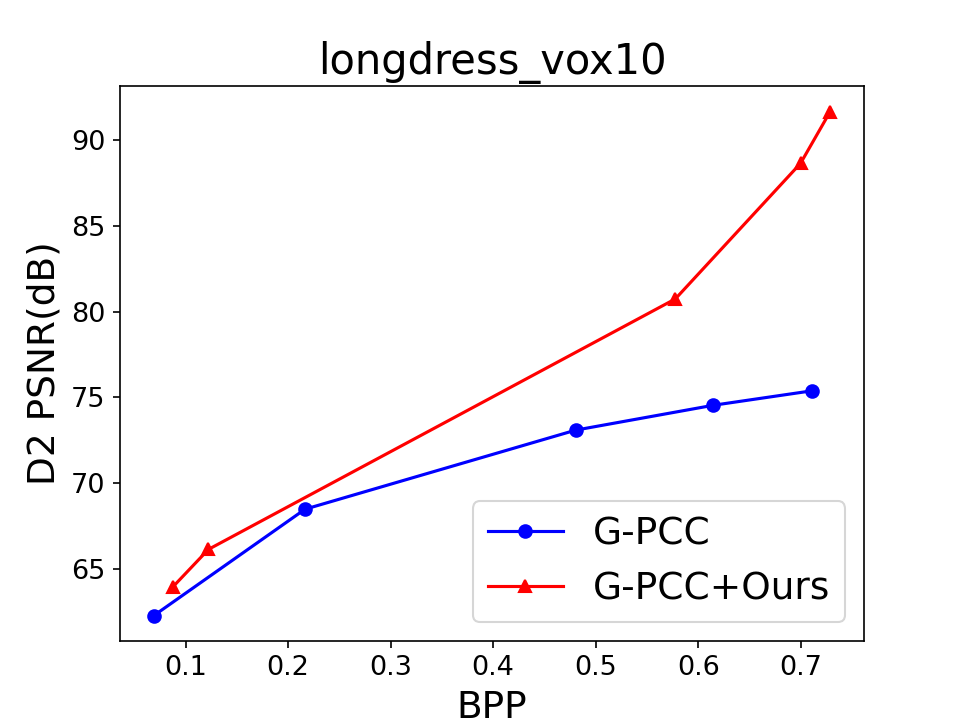}
        \includegraphics[width=0.25\linewidth]{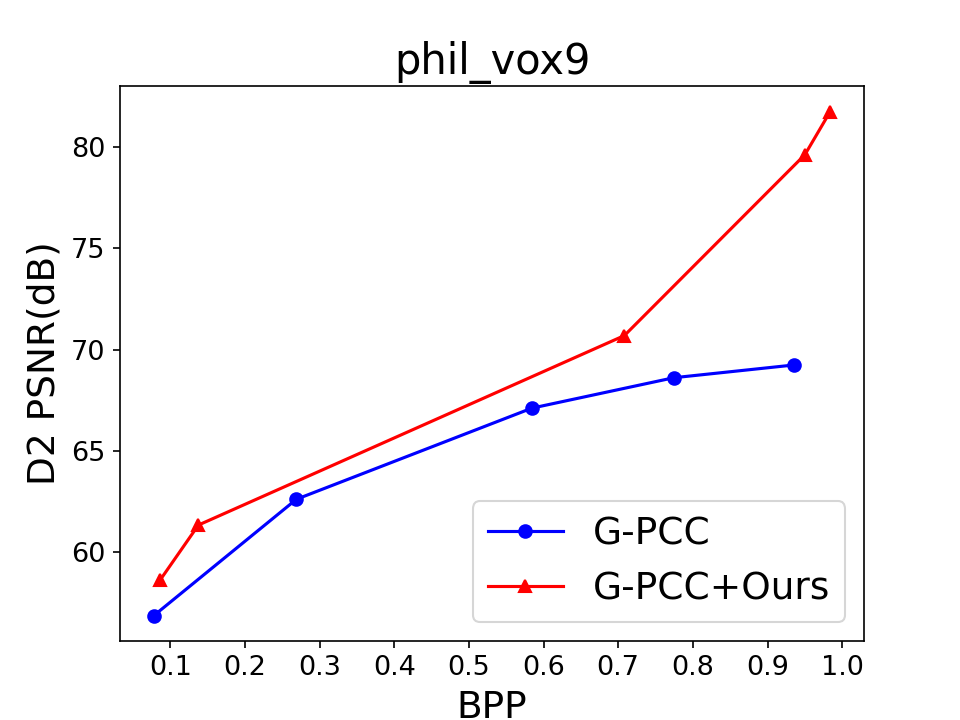}
   }
    \caption{RD curves of G-PCC with and without the proposed voxelization network.}
    \label{fig:rd_curve_gpcc}
\end{figure*}

\begin{figure*}[!t]
    \centering
    \subfigure[D1 PSNR]{
         \includegraphics[width=0.25\linewidth]{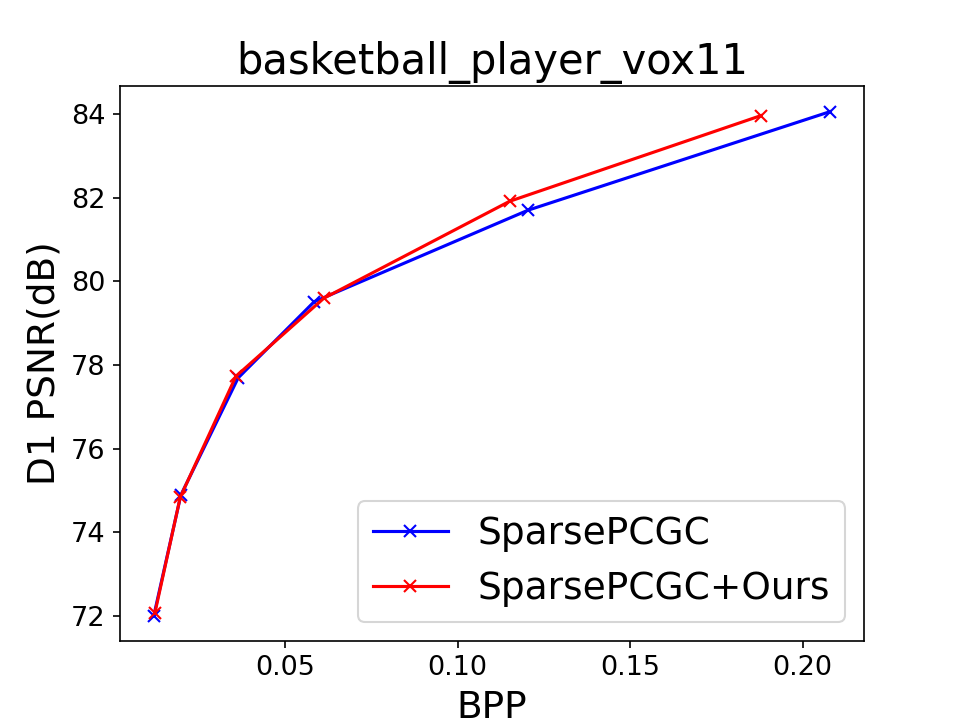}
         \includegraphics[width=0.25\linewidth]{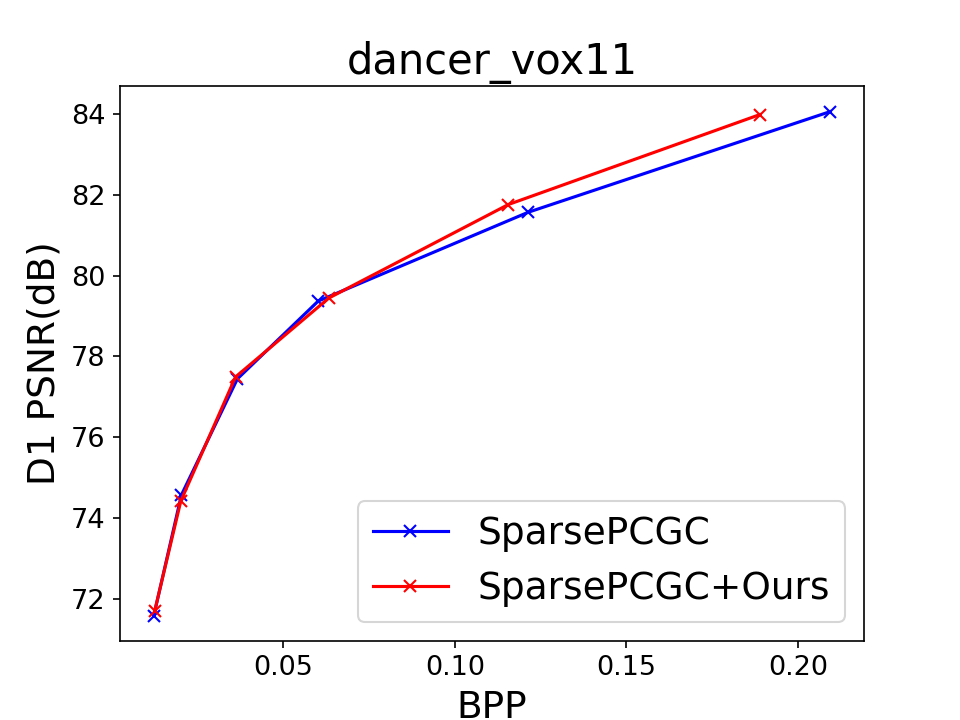}
         \includegraphics[width=0.25\linewidth]{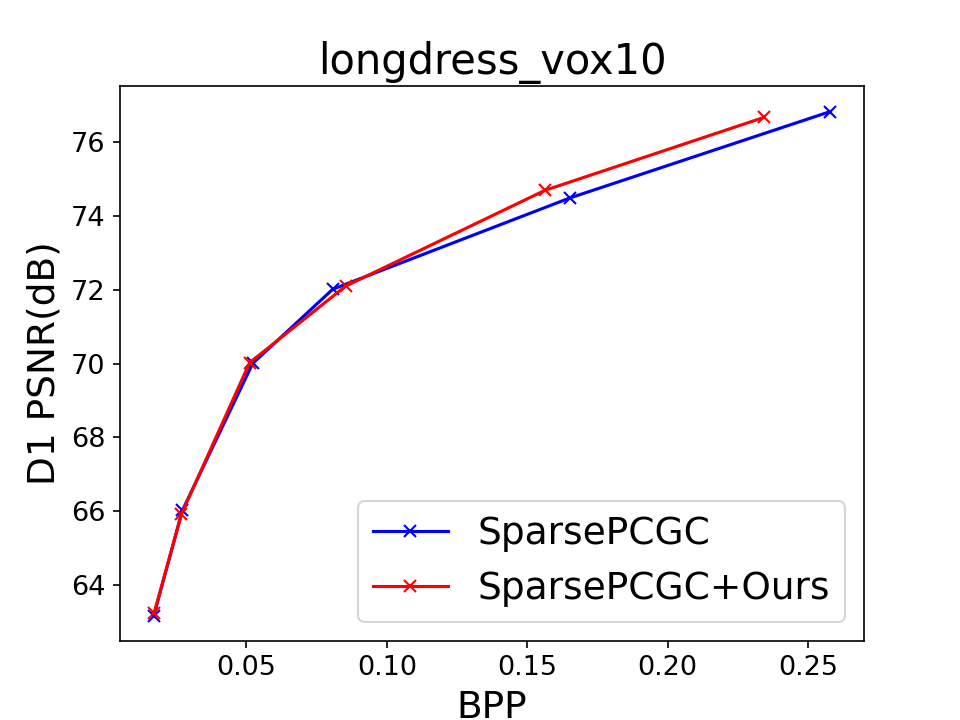}
         \includegraphics[width=0.25\linewidth]{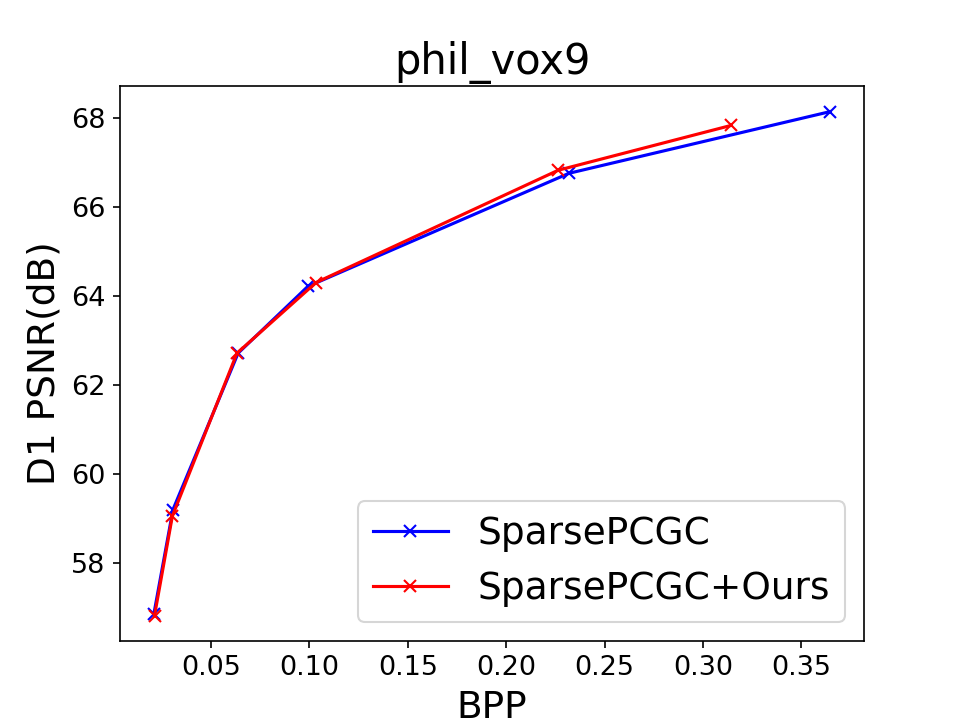}
    }
    \subfigure[D2 PSNR]{
        \includegraphics[width=0.25\linewidth]{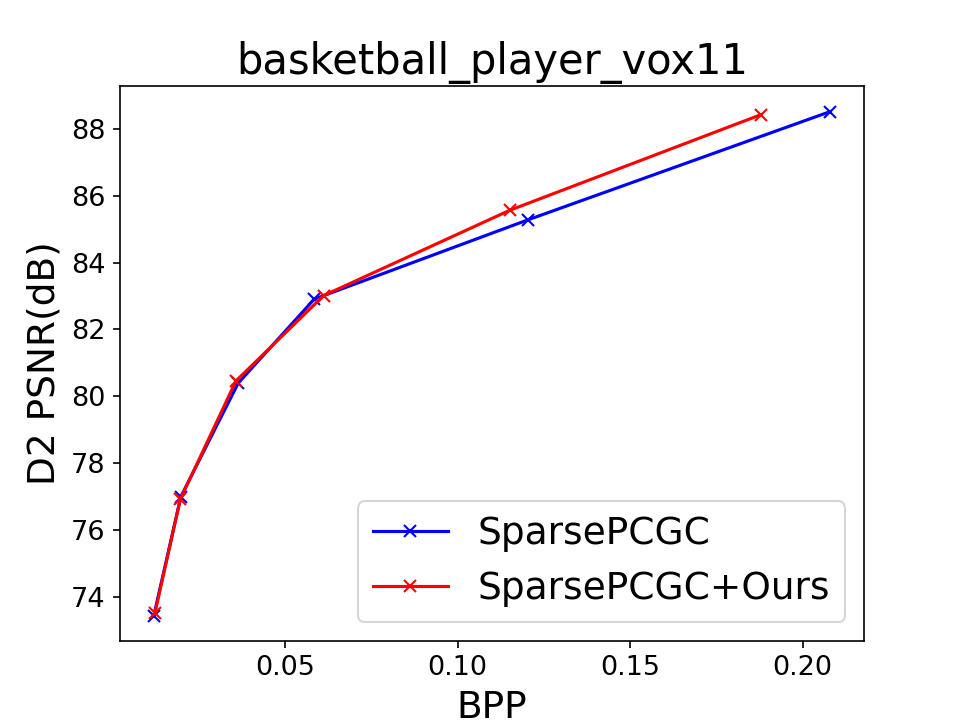}
         \includegraphics[width=0.25\linewidth]{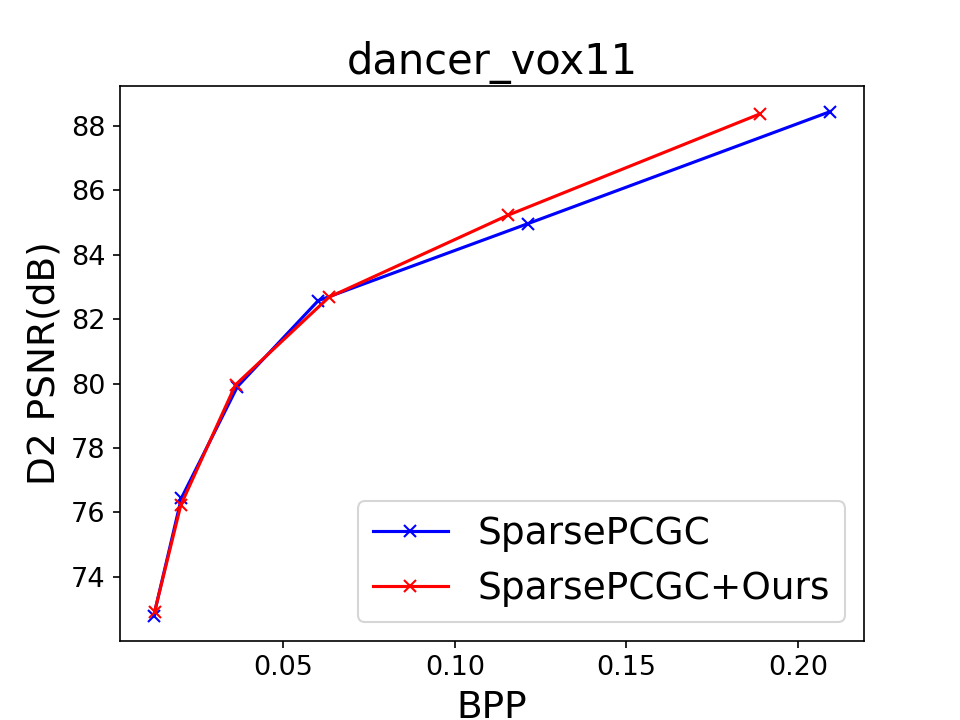}
         \includegraphics[width=0.25\linewidth]{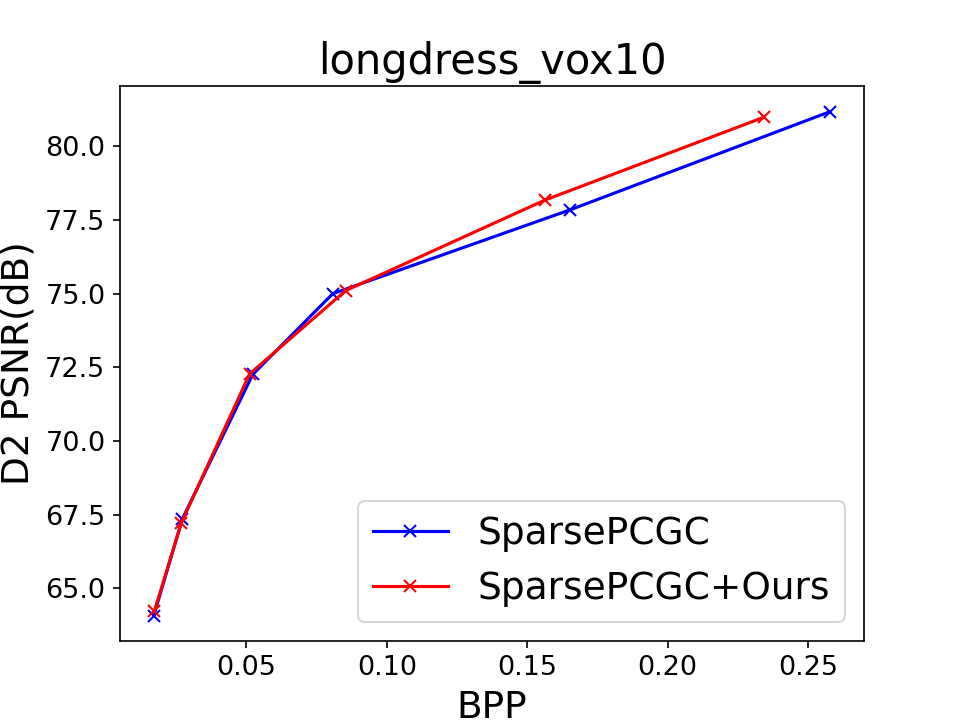}
         \includegraphics[width=0.25\linewidth]{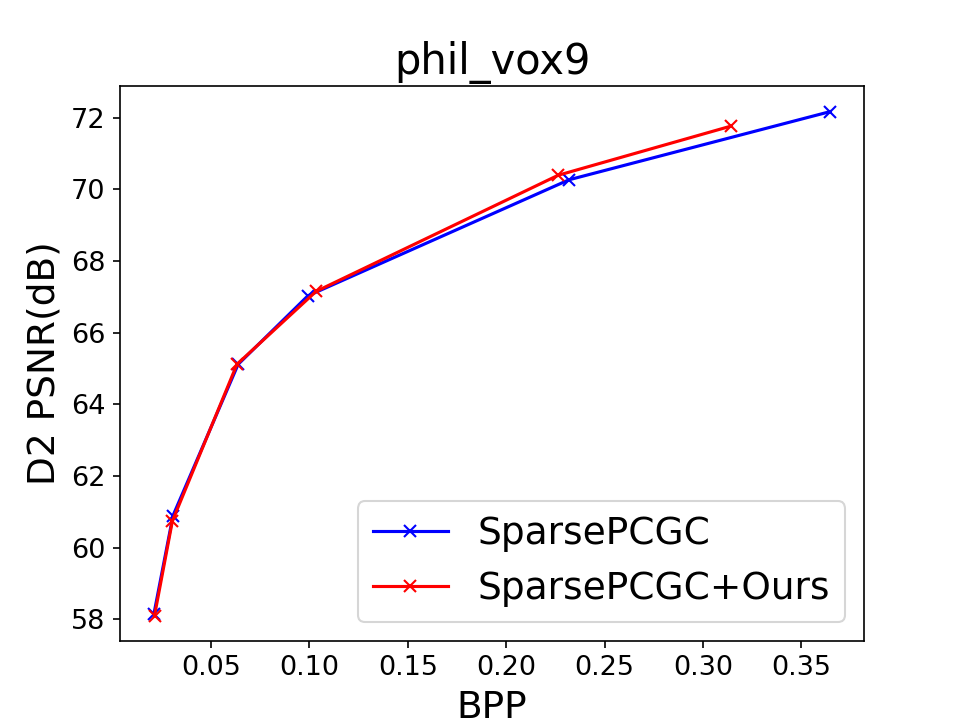}
   }
    \caption{RD curves of SparsePCGC with and without the proposed voxelization network.}
    \label{fig:rd_curve_sparsepcgc}
\end{figure*}

\begin{figure*}[!t]
    \centering
    \includegraphics[width=.9\linewidth]{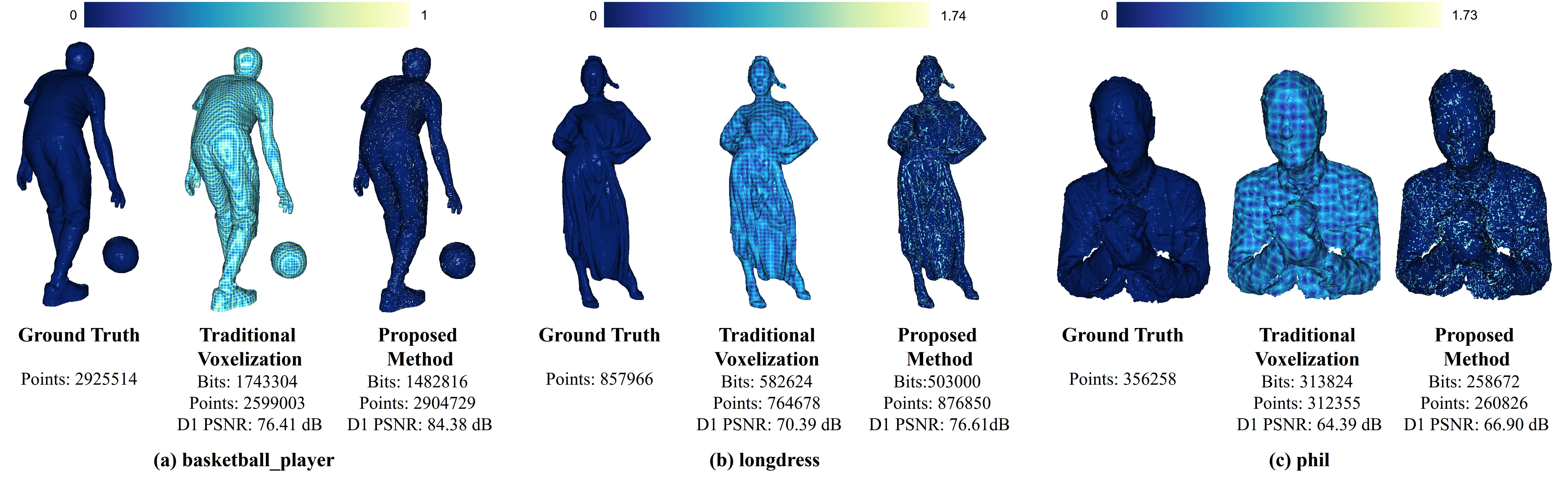}
    \caption{Visualizations of the Chamfer Distance (CD) for traditional voxelization and the proposed method. The three subfigures from left to right are the original point cloud without voxelization, the point cloud processed with traditional voxelization, and the point cloud processed with our voxelization network. The colour of the point cloud indicates the magnitude of the error.}
    \label{fig:point_cloud_diff}
\end{figure*}

\begin{figure*}[!t]
    \centering
    \includegraphics[width=.9\linewidth]{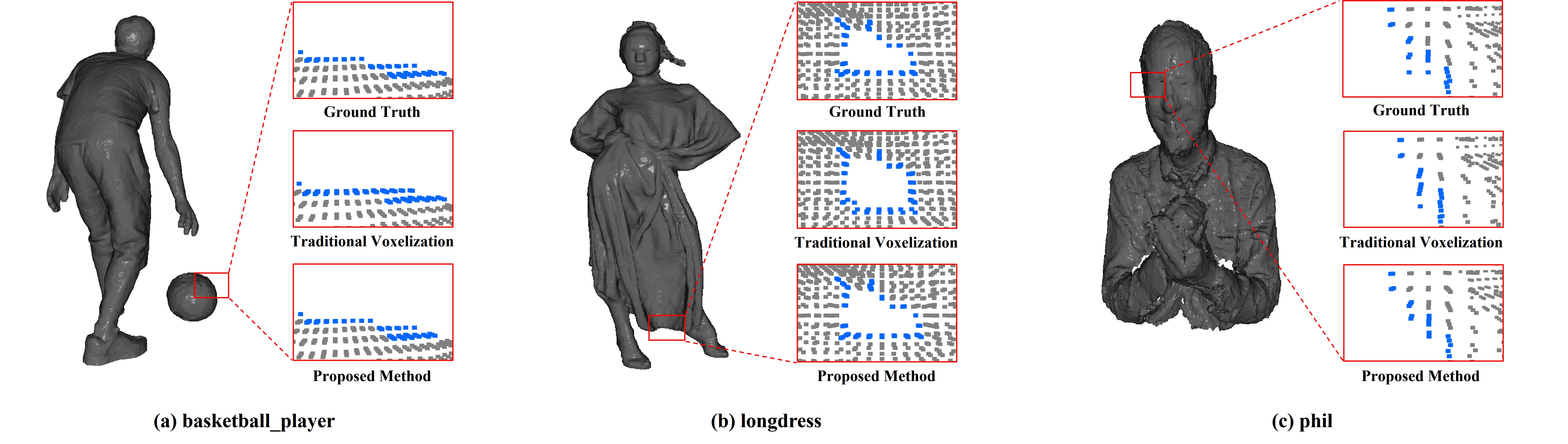}
    \caption{Visualizations of the local details for traditional voxelization and our voxelization network. Points are highlighted in blue for convenient comparison.}
    \label{fig:point_cloud_detail}
\end{figure*}

\begin{table*}[!t]
    \caption{Voxelization Time, Encoding Time, and Decoding Time.}
    \label{tab:complexity}
    \centering
    \renewcommand{\arraystretch}{1.1}
\begin{tabular}{cc|cc|c|c|cc|cc|cc}
\toprule
\multirow{3}[4]{*}{Dataset} & \multirow{3}[4]{*}{Point Cloud} & \multicolumn{4}{c|}{G-PCC \cite{gpcc}} & \multicolumn{6}{c}{SparsePCGC \cite{sparsepcgc}} \\
\cmidrule{3-12}  &   & \multicolumn{2}{c|}{Vox Time (s)} & Enc Time (s) & Dec Time (s) & \multicolumn{2}{c|}{Vox Time (s)} & \multicolumn{2}{c|}{Enc Time (s)} & \multicolumn{2}{c}{Dec Time (s)} \\
  &   & CPU & GPU & CPU & CPU & CPU & GPU & CPU & GPU & CPU & GPU \\
\midrule
\multirow{4}[1]{*}{Owlii} & basketball\_player & 11.471  & 0.112  & 6.924 & 1.539 & 11.471 & 0.112 & 100.866  & 0.185  & 134.952  & 0.607  \\
  & dancer & 9.866  & 0.096  & 5.901 & 1.352 & 9.866 & 0.096 & 147.995  & 0.179  & 113.701  & 0.553  \\
  & exercise & 8.526  & 0.088  & 4.957 & 1.213 & 8.526 & 0.088 & 80.034  & 0.176  & 91.684  & 0.513  \\
  & model & 8.750  & 0.092  & 5.034 & 1.246 & 8.750 & 0.092 & 81.011  & 0.179  & 98.599  & 0.513  \\
\multirow{4}[1]{*}{8iVFB} & longdress & 2.436  & 0.026  & 1.718 & 0.464 & 2.436 & 0.026 & 38.141  & 0.159  & 36.175  & 0.265  \\
  & loot & 2.114  & 0.023  & 1.602 & 0.432 & 2.114 & 0.023 & 34.908  & 0.157  & 31.767  & 0.245  \\
  & redandblack & 1.998  & 0.022  & 1.552 & 0.423 & 1.998 & 0.022 & 35.158  & 0.158  & 35.199  & 0.239  \\
  & soldier & 3.220  & 0.039  & 2.195 & 0.589 & 3.220 & 0.039 & 46.724  & 0.159  & 41.751  & 0.296  \\
\midrule
\multirow{4}[2]{*}{MVUB} & andrew & 0.508  & 0.009  & 0.526 & 0.148 & 0.508 & 0.009 & 15.421  & 0.246  & 12.091  & 0.174  \\
  & david & 0.595  & 0.010  & 0.625 & 0.175 & 0.595 & 0.010 & 15.927  & 0.151  & 12.956  & 0.182  \\
  & phil & 0.655  & 0.009  & 0.675 & 0.191 & 0.655 & 0.009 & 17.796  & 0.152  & 14.189  & 0.187  \\
  & sarah & 0.501  & 0.008  & 0.566 & 0.154 & 0.501 & 0.008 & 17.022  & 0.148  & 13.427  & 0.166  \\
\midrule
  & queen & 2.620  & 0.023  & 2.045 & 0.475 & 2.620 & 0.023 & 42.142  & 0.159  & 40.571  & 0.268  \\
\midrule
  & Average & 4.097  & 0.043  & 2.640 & 0.646 & 4.097 & 0.043 & 51.780  & 0.170  & 52.082  & 0.324  \\
\bottomrule
\end{tabular}%

\end{table*}

\begin{table*}[!t]
    \centering
    \caption{Performance comparison between our G-PCC surrogate model and its counterparts in terms of bits per point (bpp).}
    \begin{threeparttable}
    \renewcommand{\arraystretch}{1.1}
\begin{tabular}{r|c|c|c|c|c|c}
\toprule
\multicolumn{1}{c|}{Dataset} & Point Cloud & V-PCC \cite{gpcc} & G-PCC (octree) \cite{gpcc} & SparsePCGC \cite{sparsepcgc} \tnote{$\dagger$} & Unicorn \cite{unicorn} \tnote{$\dagger$} & Ours \\
\midrule
\multicolumn{1}{c|}{\multirow{4}[2]{*}{Owlii}} & basketball\_player & 1.113  & 0.646  & 0.520  & \text{-} & 0.464  \\
  & dancer & 1.096  & 0.629  & 0.514  & \text{-} & 0.458  \\
  & exercise & 1.034  & 0.601  & \text{-} & \text{-} & 0.445  \\
  & model & 1.013  & 0.576  & \text{-} & \text{-} & 0.454  \\
\midrule
\multicolumn{1}{c|}{\multirow{4}[2]{*}{8iVFB}} & longdress & 1.374  & 0.740  & 0.625  & 0.629  & 0.593  \\
  & loot & 1.314  & 0.702  & 0.596  & 0.600  & 0.562  \\
  & redandblack & 1.584  & 0.823  & 0.690  & 0.701  & 0.668  \\
  & soldier & 1.453  & 0.737  & 0.628  & 0.637  & 0.600  \\
\midrule
\multicolumn{1}{c|}{\multirow{4}[2]{*}{MVUB}} & andrew & 1.603  & 0.978  & \text{-} & \text{-} & 0.950  \\
  & david & 1.505  & 0.939  & \text{-} & \text{-} & 0.917  \\
  & phil & 1.715  & 0.994  & \text{-} & \text{-} & 0.964  \\
  & sarah & 1.549  & 0.928  & \text{-} & \text{-} & 0.893  \\
\midrule
  & queen & 1.006  & 0.604  & 0.550  & \text{-} & 0.518  \\
\bottomrule
\end{tabular}%

    \begin{tablenotes}
            \footnotesize
            \item[$\dagger$] These data are sourced from the cited paper.
        \end{tablenotes}
    \end{threeparttable}
    
    \label{tab:lossless_bpp}
\end{table*}

\begin{table*}[!t]
    \centering
    \caption{Performance comparison between our G-PCC surrogate model and its counterparts in terms of Compression Ratio (CR) Gain. The compression ratio is measured by the bpp gain over each counterpart.}
    \begin{threeparttable}
    \renewcommand{\arraystretch}{1.1}
\begin{tabular}{c|c|c|c|c|c}
\toprule
Dataset & Point Cloud & V-PCC \cite{gpcc} & G-PCC (octree) \cite{gpcc} & SparsePCGC \cite{sparsepcgc} \tnote{$\dagger$} & Unicorn \cite{unicorn} \tnote{$\dagger$} \\
\midrule
\multirow{4}[2]{*}{Owlii} & basketball\_player & -58.34\% & -28.25\% & -10.86\% & \text{-} \\
  & dancer & -58.17\% & -27.17\% & -10.82\% & \text{-} \\
  & exercise & -56.95\% & -25.88\% & \text{-} & \text{-} \\
  & model & -55.14\% & -21.14\% & \text{-} & \text{-} \\
\midrule
\multirow{4}[2]{*}{8iVFB} & longdress & -56.84\% & -19.87\% & -5.10\% & -5.70\% \\
  & loot & -57.18\% & -19.90\% & -5.63\% & -6.26\% \\
  & redandblack & -57.84\% & -18.88\% & -3.22\% & -4.74\% \\
  & soldier & -58.72\% & -18.67\% & -4.51\% & -5.86\% \\
\midrule
\multirow{4}[2]{*}{MVUB} & andrew & -40.71\% & -2.89\% & \text{-} & \text{-} \\
  & david & -39.11\% & -2.43\% & \text{-} & \text{-} \\
  & phil & -43.82\% & -3.10\% & \text{-} & \text{-} \\
  & sarah & -42.34\% & -3.73\% & \text{-} & \text{-} \\
\midrule
  & queen & -48.51\% & -14.22\% & -5.86\% & \text{-} \\
\bottomrule
\end{tabular}%
    
    \begin{tablenotes}
        \footnotesize
        \item[$\dagger$] These data are sourced from the cited papers.
    \end{tablenotes}
    \end{threeparttable}
    \label{tab:lossless_cr}
\end{table*}

\section{Experiments}
\subsection{Implementation Details}
For training, we use the ShapeNet dataset\cite{shapenet}. We densely sample points on raw meshes to generate point clouds, and then randomly rotate and quantize them with 8-bit geometry precision at each dimension. For testing, we select three widely adopted point cloud datasets: Microsoft Voxelized Upper Bodies(MVUB) including  \textit{andrew\_frame0000}, \textit{david\_frame0000}, \textit{phil\_frame0139}, \textit{sarah\_frame0023}, 8i Voxelized Full Bodies (8iVFB) including \textit{longdress\_1300}, \textit{loot\_1200}, \textit{redandblack\_vox10\_1550}, \textit{soldier\_0690}, Owlii Dynamic Human Textured Mesh Sequence Dataset (Owlii) including \textit{basketball\_player\_00000200}, \textit{dancer\_00000001}, \textit{exercise\_00000001}, \textit{model\_00000001}. Owlli, 8iVFB, and MVUB were set to 11-bit, 10-bit, and 9-bit to adequately cover the point clouds with different bit depth precisions. Additionally, we included the 10-bit point cloud queen (\textit{queen\_0200}) as an extra test case. Some samples of the training and testing datasets  are visualized in Fig.\ref{fig:data_vis} for an  intuitive observation. The distortions were measured using the point-to-point PSNR (D1 PSNR) and the point-to-plane PSNR (D2 PSNR)~\cite{ctc_gpcc}. Rate-distortion efficiency was quantified using the Bjøntegaard-Delta (BD) rate metric.

For joint optimization with the G-PCC surrogate model, the Adam optimizer was adopted with a batch size of 8. We first trained the surrogate model for 10 epochs. Then, the parameters of the surrogate model were fixed and the voxelization network was trained for 30 epochs. The initial learning rate was set to 0.0001, decaying by a factor of 0.5 every 5 epochs.

To further prove the generalization of the proposed voxelization network for learning-based codecs, SparsePCGC was chosen to be jointly optimized with our voxelization method. As SparsePCGC accepts point clouds already pruned by the~\textit{MinkowskiPruning} function in the Minkowski Engine~\cite{me}, we thus slightly modify SparsePCGC to accept point clouds without pruning. All other experimental configurations were the same as described in the original paper.

All experiments were conducted on a computer equipped with the 13th Gen Intel(R) Core(TM) i5-13600KF CPU and the Nvidia RTX 3090 24 GB GPU.

\subsection{Versatile Voxelization Network as Preprocessing Module}
This subsection evaluates the performance gain achieved by integrating the proposed voxelization network as a preprocessing module for G-PCC~\cite{gpcc} (TMC13-v23) and SparsePCGC). 

As presented in Table~\ref{tab:bd_rate}, integrating our voxelization network with G-PCC achieves an average BD-Rate gain of $38.84\%$ in D1 PSNR and $27.90\%$ in D2 PSNR. When combined with SparsePCGC, it achieves an average BD-Rate gain of $2.24\%$ in D1 PSNR and $2.90\%$ in D2 PSNR. Rate-distortion curves for individual point clouds are shown in Fig.~\ref{fig:rd_curve_gpcc} and Fig.~\ref{fig:rd_curve_sparsepcgc}. These results demonstrate that our voxelization network consistently improves the rate-distortion performance of both G-PCC and SparsePCGC across datasets with different precision levels.

To provide a visual comparison of the proposed voxelization network against the widely-used voxelization in G-PCC~\cite{voxelization_hinks}, Figure~\ref{fig:point_cloud_diff} illustrates the comparison of the Chamfer Distance (CD) for both approaches. The colour of the point cloud indicates the magnitude of the error. Objective D1 PSNR, number of points, and the G-PCC-compressed bit size are also provided. The proposed voxelization introduces smaller errors, resulting in higher PSNR. Figure~\ref{fig:point_cloud_detail} further highlights localized details of the voxelized point clouds. These details demonstrate that our approach better preserves structural features compared to traditional voxelization.

Additionally, while traditional voxelization can only reduce the number of points, our voxelization network adaptively adjusts the point density—either reducing or increasing it—to optimize coding performance. As shown in Fig.~\ref{fig:point_cloud_diff} (a) and Fig.~\ref{fig:point_cloud_diff} (b), our method produces point clouds with more points than traditional voxelization while requiring fewer bits for encoding, further underscoring its efficiency.  

\begin{figure}[!t]
    \centering
    \includegraphics[width=\linewidth]{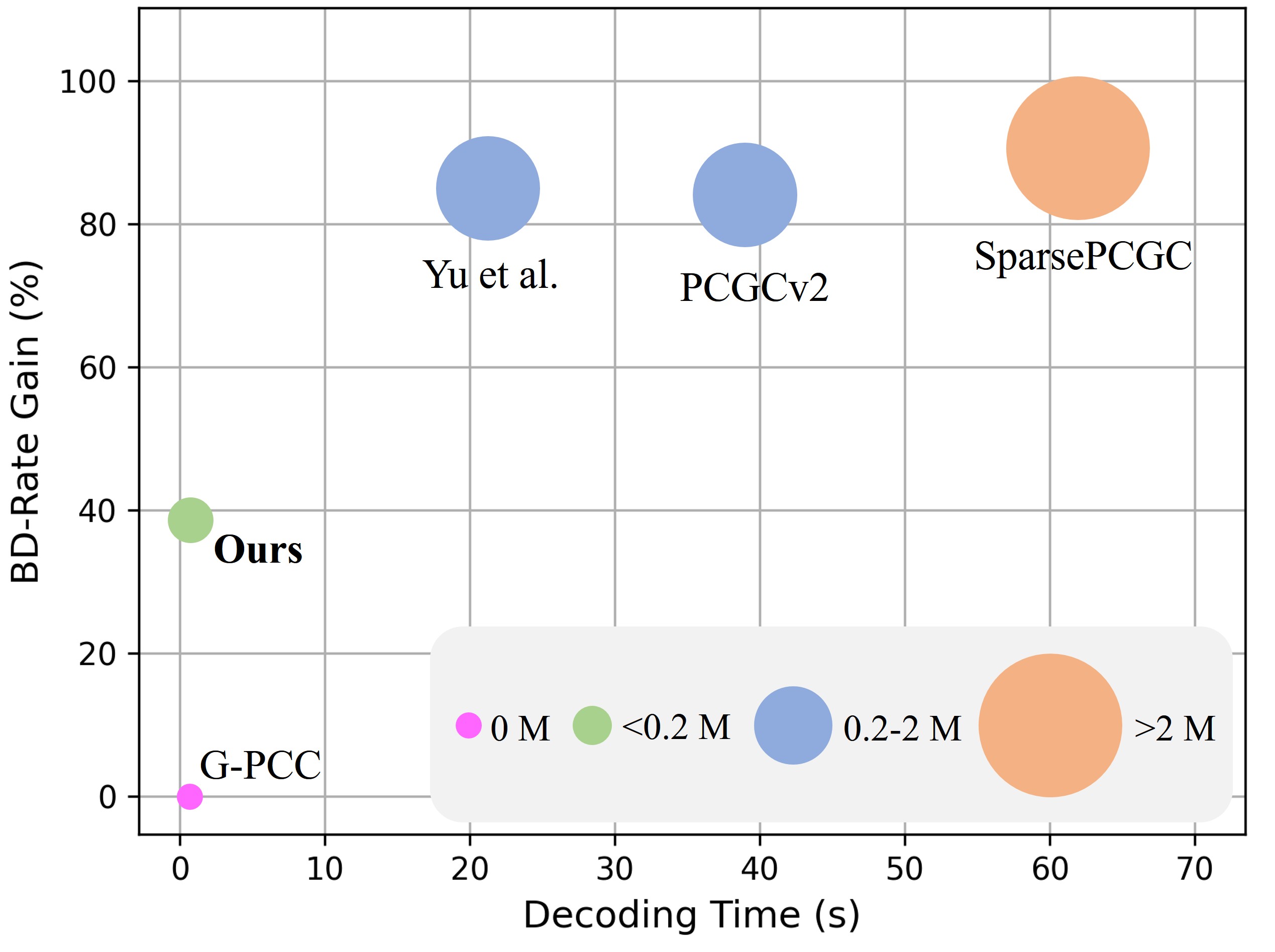}
    \caption{Decoding Time (s) vs. BD-Rate Gain (\%) vs. Parameters (M). The BD-rate gain is calculated using the G-PCC as the anchor.}
    \label{fig:rel_perf}
\end{figure}

Finally, the comparison between our voxelization followed by G-PCC with other codecs is shown in Fig.~\ref{fig:rel_perf}. Our voxelization network improves the rate-distortion performance of G-PCC, with slight extra parameters but without extra decoding time. Although not as good as other learning-based methods (Yu et al., PCGCv2, SparsePCGC) in terms of BD-rate gain, our method has a large lead in the number of parameters and decoding time. Moreover, to highlight the lightweight nature of our voxelization network, we report the voxelization time, G-PCC encoding/decoding time, and SparsePCGC encoding/decoding time in Table~\ref{tab:complexity}. When deploying neural networks, minimizing hardware costs is crucial for widespread adoption. Considering that CPUs are generally less expensive than GPUs, many studies \cite{mobilenetv4, imdn, shufflenet,efficient_net, fbnet, masnet, AIM19constrainedSR, mobilenetv3, mobilenetv2} use CPU platforms as benchmarks for lightweight model development. This paper adopts the same comparative strategy. The results demonstrate that the voxelization time is competitive compared to the encoding and decoding times, underscoring the efficiency of our approach.

\subsection{G-PCC Surrogate as Standalone Lossless Codec}
Our G-PCC surrogate model is capable of predicting the occupancy probabilities of nodes in the octree. This enables its direct application as a lossless point cloud codec. In this section, we compare the performance of our surrogate model against several state-of-the-art methods including V-PCC (TMC2-v24), G-PCC (TMC13-v23), OctAttention \cite{octattn}, SparsePCGC \cite{sparsepcgc} and Unicorn \cite{unicorn}.

The performance comparison is presented in Table \ref{tab:lossless_bpp} and \ref{tab:lossless_cr}. In Table \ref{tab:lossless_cr}, we show the Compression Ratio (CR) gain of our surrogate model. The CR gain is measured by the bpp gain over each counterpart. A negative CR gain indicates superior performance compared to the anchor method. The results demonstrate that our surrogate model is highly competitive as a lossless codec.

\section{Conclusion}
This paper proposes a novel solution for point cloud compression by enhancing traditional and learning-based PCC codec with a voxelization network that preprocesses the point cloud in a jointly optimized manner. To enable end-to-end optimization with a two-sided gradient propagation pathway, we introduced a differentiable G-PCC surrogate model, which also serves as an effective codec for lossless point cloud compression. Our approach achieves significant improvements in rate-distortion efficiency, demonstrating competitive performance against state-of-the-art methods.

A key advantage of the proposed framework is its user-friendly design, which requires no modifications on the user side and introduces no additional computational burden for end users. The proposed framework brings no modification on the user side, thus introducing no extra burden for end users. By combining the high compression performance of jointly optimized learning-based methods with the guaranteed reproducibility and computational efficiency of traditional G-PCC, our work paves the way for practical and deployable PCC codecs suitable for real-world point cloud applications.

\bibliographystyle{IEEEtran}
\bibliography{myref.bib}

\end{document}